\documentclass{article} %
\usepackage{colm2024_conference}

\usepackage{booktabs}
\usepackage{graphicx}
\usepackage{enumitem}
\usepackage{wrapfig}
\usepackage{algorithm}
\usepackage{algpseudocode}
\usepackage{wrapfig}
\usepackage{float}
\usepackage{microtype}
\usepackage{amsmath}
\usepackage{amssymb}
\usepackage{colortbl}
\usepackage{fontawesome5}
\usepackage[utf8]{inputenc}
\definecolor{lightgray}{rgb}{0.9,0.9,0.9}
\usepackage{caption}
\usepackage{subcaption}
\usepackage{xcolor}
\usepackage[table]{xcolor}
\usepackage{setspace}
\usepackage{url}
\usepackage{multirow}
\usepackage{colortbl}
\usepackage{tabularx}
\usepackage{blindtext}
\usepackage{pgfplots}
\pgfplotsset{compat=1.18} 
\usepackage{tikz}
\usetikzlibrary{er,positioning,bayesnet}
\usepackage{makecell}
\usepackage{tipa}
\usepackage{siunitx}
\usepackage{nicefrac}
\usepackage{tocloft}
\usepackage{listings}
\usepackage[raster,skins]{tcolorbox} %
\usepackage{xltabular}
\usepackage{adjustbox}
\usepackage{xurl}
\usepackage{multicol}
\usepackage{xcolor}

\usepackage{amsmath,amsfonts,bm}

\def\eqref#1{equation~\ref{#1}}
\def\1{\bm{1}}

\DeclareMathAlphabet{\mathsfit}{\encodingdefault}{\sfdefault}{m}{sl}
\SetMathAlphabet{\mathsfit}{bold}{\encodingdefault}{\sfdefault}{bx}{n}

\makeatletter
\DeclareRobustCommand\onedot{\futurelet\@let@token\@onedot}
\def\@onedot{\ifx\@let@token.\else.\null\fi\xspace}

\def\eg{\emph{e.g}\onedot}

\makeatother

\definecolor{GEPrompt}{HTML}{ACB6F3} 
\usepackage{tcolorbox}
\tcbuselibrary{skins,breakable,listingsutf8}
\tcbset{examplebox/.style={
  colback=GEPrompt!8!white, colframe=GEPrompt!75!black, coltitle=GEPrompt!50!black,
  fonttitle=\bfseries, boxrule=0.5mm, sharp corners, enhanced,
  left=2mm,right=2mm,top=2mm,bottom=2mm,
  attach boxed title to top left={yshift=-2mm,xshift=5mm},
  boxed title style={colframe=GEPrompt!75!black,colback=white,sharp corners}}}
\newtcblisting{promptlisting}[2][]{%
  examplebox, breakable, listing only, title=~#2,
  listing options={basicstyle=\ttfamily\small,breaklines=true, breakatwhitespace=false,columns=fullflexible,keepspaces=true, showstringspaces=false,tabsize=2},
  #1}

\definecolor{FirstPlace}{HTML}{ACB6F3}
\definecolor{SecondPlace}{HTML}{F4BB6E}

\title{OmniVBench: A Benchmark and Large-Scale Dataset for \\ Omni Reference-to-Video Generation}

\author{Wenxue Li$^{3*}$ \quad Peiyan Guan$^{1*}$ \quad Haoyang Jiang \quad  Junxian Cai$^{1}$ \quad Hualuo Liu$^{1}$ \quad Chunjie Zhang$^{1}$ \quad Chong Guan$^{1}$ \quad Kai Huang$^{1}$ \quad Songlian Li$^{1}$ \\ Taiyi Wu$^{1}$ \quad Yongjian Yu$^{1}$ \quad Xiaotong Zhao$^{1}$ \quad Alan Zhao$^{1}$ \\ Quande Liu$^{2,\dagger}$ \quad Xi Chen$^{1,\dagger}$ \quad Yu Liu$^{1}$ \quad Lei Zhu$^{3,\dagger}$ \\[4pt] $^{1}$\textbf{Online Video BU, Tencent} \\ $^{2}$\textbf{The Chinese University of Hong Kong} \\ $^{3}$\textbf{The Hong Kong University of Science and Technology (Guangzhou)} \\ [2pt] {\qquad \small $^*$Equal Contribution. } \small $^\dagger$Corresponding Author.  }

\begin{document}

\maketitle

\begin{figure}[h]
\centering
\includegraphics[width=1\linewidth]{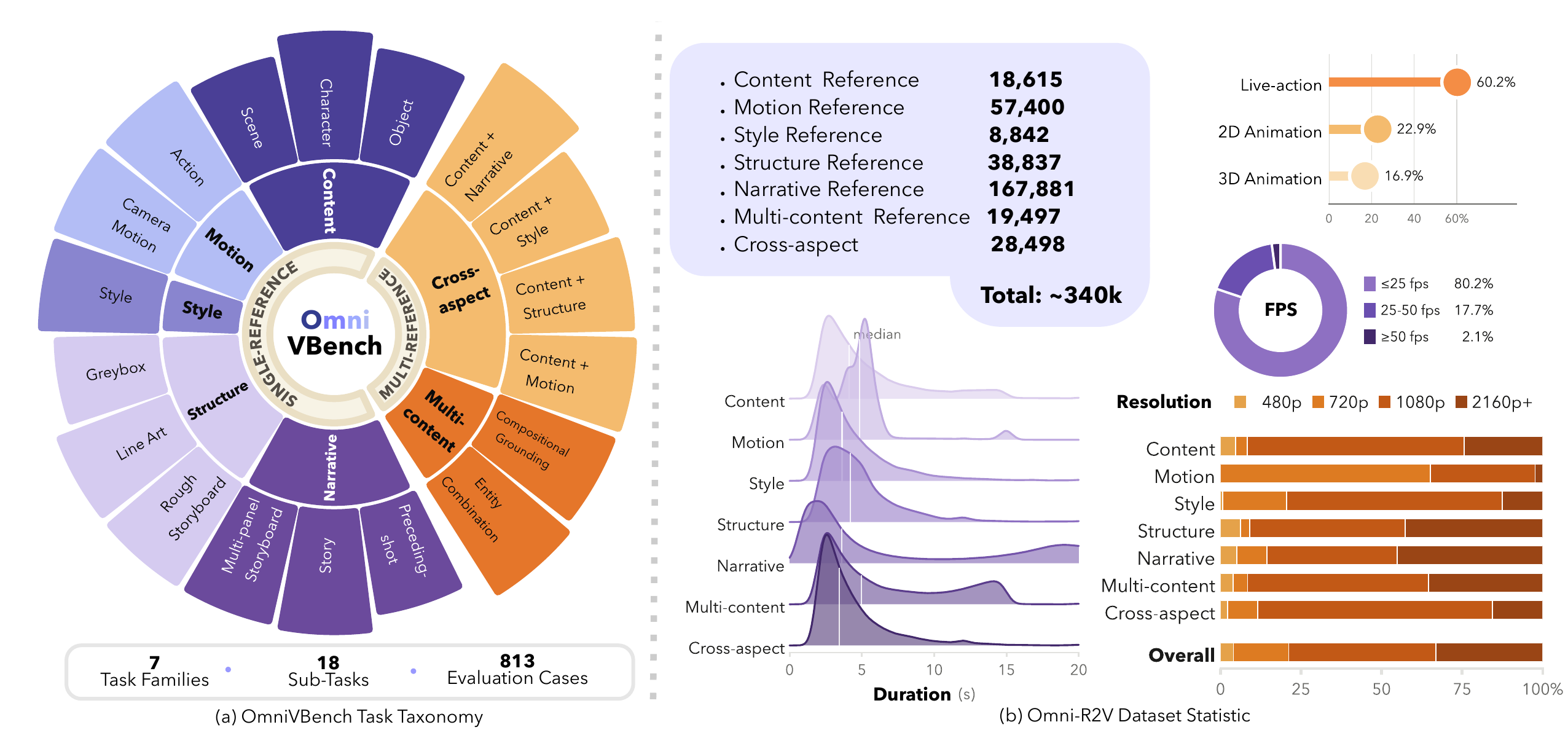}
\vspace{-10pt}
\caption{
Overview of \textbf{OmniVBench} and the \textbf{Omni-R2V Dataset}.}

\label{fig:teaser-top}
\end{figure}

\begin{abstract}
Omni reference-to-video (R2V) generation is emerging as a general paradigm that enables flexible video customization from diverse and compositional visual references.
However, existing benchmarks fall short of these emerging capabilities: their test cases cover only a limited range of reference types and compositions, and their evaluation protocols largely assess holistic reference consistency, overlooking whether reference factors are properly preserved, disentangled, and routed.
Meanwhile, the high cost of constructing omni R2V training data makes suitable training resources scarce.
To address these gaps, we introduce \textbf{OmniVBench} and the \textbf{Omni-R2V Dataset}, establishing a shared foundation for evaluating and training omni R2V models.
\textbf{OmniVBench} expands R2V evaluation across broader reference types, richer reference compositions, and more flexible instruction-driven customization, covering 7 task families and 18 fine-grained tasks spanning content, motion, style, structure, narrative, and multi-reference settings.
Complementing this task design, we introduce factor-grounded evaluation with 12,172 case-specific checklist items, explicitly assessing whether intended reference factors are faithfully preserved, correctly disentangled and bound to their targets, and realized according to the instruction.
We further introduce the \textbf{Omni-R2V Dataset}, bringing industrial-grade training resources for diverse R2V tasks to the broader research community. Drawing primarily on a large-scale corpus of professional video footage, it comprises 340K processed training samples spanning diverse reference types and multi-reference compositions.
Specifically, we develop task-specific pipelines for reference-target pair construction, offering a practical and scalable recipe for omni R2V data construction.
Extensive evaluation of advanced open- and closed-source R2V models reveals substantial performance variation across diverse tasks and distinct capability patterns across evaluation dimensions, demonstrating that OmniVBench remains discriminative and diagnostically informative even for frontier R2V models.
\end{abstract}

\section{Introduction}
Recent advances in generative models and multimodal large language models have substantially expanded the controllability of video generation, enabling reference-to-video (R2V) generation to incorporate visual references as flexible control signals for video synthesis. 
R2V is rapidly evolving from specialized generation with isolated references~\citep{Lynx,Stand-In,Slot-ID} toward more general settings involving compositional references~\citep{VideoAlchemist,VACE,hunyuancustom,univideo,HuMo,omnishow,OmniVCus,OmniWeaving_IntelligentVBench,bernini,loomvideo,Vino,Dreamid-omni_bytedance,PoCo,harmoview}.
As R2V moves toward practical creative workflows, reference control is becoming increasingly diverse, compositional, and flexible, giving rise to the more general paradigm of \textit{omni R2V generation}.

However, this broader R2V paradigm poses new challenges for systematic evaluation. 
While several benchmarks~\citep{OmniWeaving_IntelligentVBench,UniVBench,OpenS2V-Nexus,MultiRef-Compass} have been developed to assess R2V models, they remain limited in both evaluation coverage and granularity: 
\textbf{\textit{(1) Limited task and scenario coverage.}} As summarized in Table~\ref{tab:benchmark_comparison}, existing benchmarks capture only a subset of the increasingly diverse R2V task space, with evaluation largely centered on content-oriented references. 
Broader reference types, richer reference compositions, and fine-grained instruction-driven customization therefore remain insufficiently evaluated, despite their growing importance in practical creative workflows.
\textbf{\textit{(2) Limited factor-level assessment.}} Existing benchmarks~\cite{OpenS2V-Nexus,OmniWeaving_IntelligentVBench,loomvideo} primarily evaluate overall fidelity, both in test-case design and evaluation metrics. 
However, omni R2V requires fine-grained control over how reference factors are preserved, modified, routed to specific targets, and combined across multiple references according to the instruction.
Assessing these capabilities requires factor-aware test cases and fine-grained evaluation of how each intended reference factor is utilized. 

\definecolor{OursBlue}{HTML}{ACB6F3}
\definecolor{SubHeaderBlue}{RGB}{240,246,252}
\begin{table*}[!t]
\centering
\caption{Comparison of reference task coverage between OmniVBench and existing reference-based video generation benchmarks.}
\label{tab:benchmark_comparison}
\resizebox{\textwidth}{!}{%
\begin{tabular}{lcccccccccccccc}
\toprule
\multirow{2}{*}{\textbf{Benchmark}}
& \multicolumn{3}{c}{\textbf{Content Ref.}}
& \multicolumn{2}{c}{\textbf{Motion Ref.}}
& \multicolumn{1}{c}{\textbf{Style Ref.}}
& \multicolumn{3}{c}{\textbf{Structure Ref.}}
& \multicolumn{3}{c}{\textbf{Narrative Ref.}}
& \multicolumn{2}{c}{\textbf{Multiple Refs.}} \\
\cmidrule(lr){2-4}
\cmidrule(lr){5-6}
\cmidrule(lr){7-7}
\cmidrule(lr){8-10}
\cmidrule(lr){11-13}
\cmidrule(lr){14-15}
& \makecell[c]{Object}
& \makecell[c]{Character}
& \makecell[c]{Scene}
& \makecell[c]{Action}
& \makecell[c]{Camera\\Motion}
& \makecell[c]{Style}
& \makecell[c]{Greybox}
& \makecell[c]{Line\\Art}
& \makecell[c]{Rough\\Storyboard}
& \makecell[c]{Multi-Panel\\Storyboard}
& \makecell[c]{Story}
& \makecell[c]{Preceding-\\Shot}
& \makecell[c]{Multi-\\Content}
& \makecell[c]{Cross-\\Aspect}\\
\midrule
OpenS2V-Eval~\cite{OpenS2V-Nexus}
& \checkmark & \checkmark & 
& & 
& 
& & & & 
& & 
& \checkmark & \\

VACE-Bench~\cite{VACE}
& \checkmark & \checkmark & 
& \checkmark &  & 
&  & \checkmark & 
& & & 
& \checkmark & \\

UniVBench~\cite{UniVBench}
& \checkmark & \checkmark & \checkmark
& & 
& 
& & & 
& & & 
& \checkmark & \\

IntelligentVBench~\cite{OmniWeaving_IntelligentVBench}
& \checkmark & \checkmark & \checkmark
& & 
& 
& & & 
& & & 
& \checkmark & \\

FashionVideoBench~\cite{loomvideo}
& \checkmark & \checkmark & \checkmark
& \checkmark & 
& 
& & & & 
& & 
& \checkmark \\

\midrule
\rowcolor{OursBlue}
\textbf{OmniVBench (Ours)}
& \checkmark & \checkmark & \checkmark
& \checkmark & \checkmark
& \checkmark 
& \checkmark & \checkmark & \checkmark & \checkmark
& \checkmark & \checkmark
& \checkmark & \checkmark  \\
\bottomrule
\end{tabular}%
}
\end{table*}
\definecolor{OursOrange}{HTML}{F4BB6E}
\definecolor{SubHeaderBlue}{RGB}{240,246,252}

\begin{table*}[!t]
\centering
\caption{Comparison of Omni-R2V Dataset with existing R2V datasets in terms of scale, reference modalities, reference task coverage, and processed data provision (i.e., whether processed samples are directly provided without requiring users to download and process raw source videos).}
\label{tab:dataset_comparison}

\setlength{\tabcolsep}{9pt}
\resizebox{\textwidth}{!}{%
\begin{tabular}{lccccccccccc}
\toprule
\multirow{2}{*}{\textbf{Dataset}}
& \multirow{2}{*}{\textbf{\# Size}}
& \multicolumn{2}{c}{\textbf{Reference Modality}}
& \multicolumn{7}{c}{\textbf{Reference Task Coverage}}
& \multirow{2}{*}{\makecell[c]{\textbf{Processed Data}\\ \textbf{Provided}}}  \\

\cmidrule(lr){3-4}
\cmidrule(lr){5-11}
&
& \makecell[c]{Image}
& \makecell[c]{Video}
& \makecell[c]{Content}
& \makecell[c]{Motion}
& \makecell[c]{Style}
& \makecell[c]{Structure}
& \makecell[c]{Narrative}
& \makecell[c]{Multi-\\Content}
& \makecell[c]{Cross-\\Aspect}
&  \\

\midrule

OpenS2V-5M~\citep{OpenS2V-Nexus}
& 5.4M

& \checkmark & 
& \checkmark
&
&
&
& 
& \checkmark
& 
& \checkmark\\

Phantom-Data~\cite{phantom-data}
& 1M
& \checkmark & 
& \checkmark
& 
&
&
&
& \checkmark
& & \\

MuSS~\cite{muss}
& 30K
& \checkmark & 
& \checkmark
& 
&
&
&
&
& 
& \\

\midrule

\rowcolor{OursOrange}
\textbf{Omni-R2V (Ours)}
& \textbf{340K}
& \checkmark
& \checkmark
& \checkmark
& \checkmark
& \checkmark
& \checkmark
& \checkmark
& \checkmark
& \checkmark
& \checkmark \\

\bottomrule
\end{tabular}%
}
\end{table*}

The growing diversity of R2V tasks also calls for training data that cover a broader range of reference types and their compositions. As shown in Table~\ref{tab:dataset_comparison}, existing datasets are largely tailored to specific tasks or individual reference types, leaving comprehensive R2V training resources scarce. 
However, constructing comprehensive R2V data at scale is challenging. Different reference factors require specialized construction pipelines, increasing the complexity and cost as task coverage expands. Moreover, constructing high-quality R2V pairs requires large and diverse video collections to support diverse reference–target relationships.

\begin{figure*}[t]
\centering{\includegraphics[width=0.93\textwidth]{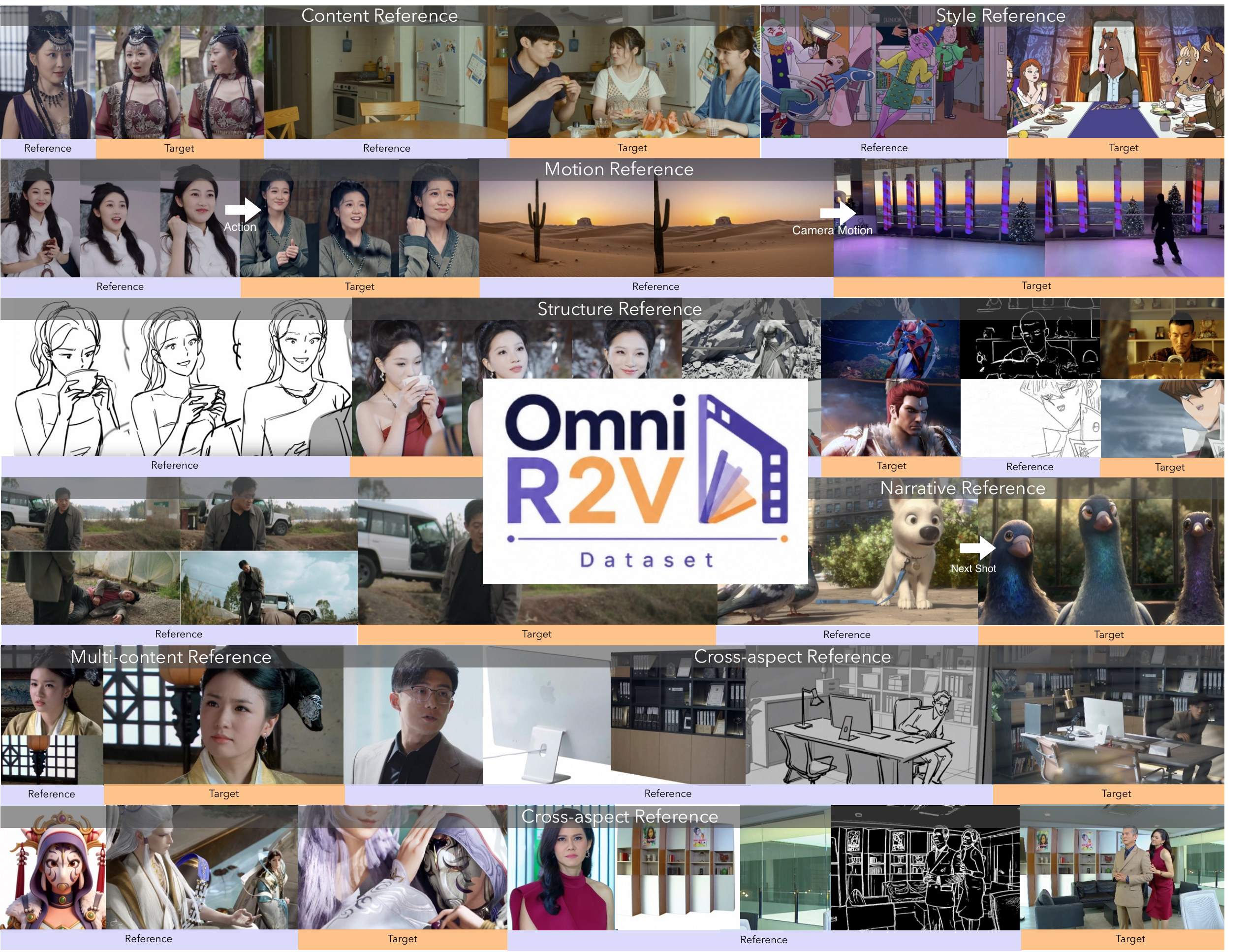}}
\caption{Representative reference--target pairs from the Omni-R2V Dataset.}
\label{Fig.data_show}
\end{figure*}

To address these challenges, we introduce \textbf{OmniVBench} and the \textbf{Omni-R2V Dataset}, providing complementary evaluation and training resources for omni R2V generation. 
\textbf{OmniVBench} systematically expands R2V evaluation across heterogeneous reference types, diverse reference compositions, and fine-grained instruction-driven customization. It organizes the R2V task space into multiple dimensions—content, motion, style, structure, and narrative, as well as their multi-reference compositions. 
Beyond broad task coverage, we introduce factor-grounded evaluation, where each case is decomposed into case-specific checklists that explicitly assess whether intended reference factors are faithfully preserved, correctly disentangled and routed to their targets, and realized according to the instruction. In total, OmniVBench comprises 12,172 factor-grounded checklist items, enabling fine-grained diagnosis beyond holistic reference consistency.
We further introduce the \textbf{Omni-R2V Dataset}, a large-scale training resource designed to support the diverse and compositional nature of omni R2V generation. Drawing primarily on a large-scale corpus of professional video footage, it comprises 340K processed R2V training samples spanning heterogeneous reference types and multi-reference compositions, with representative reference–target pairs shown in Fig.~\ref{Fig.data_show}. To support diverse R2V tasks at scale, we develop task-specific pipelines for constructing reference–target pairs and corresponding training instructions.

The main contributions of this work are:
\begin{itemize}
    \item We introduce \textbf{OmniVBench}, a comprehensive R2V benchmark spanning 7 task families and 18 fine-grained tasks across heterogeneous reference types, diverse reference compositions, and flexible instruction-driven customization. Beyond broad task coverage, we introduce a factor-grounded evaluation protocol with 12,172 case-specific checklist items, enabling fine-grained evaluation beyond holistic reference assessment.
    \item We construct and release the \textbf{Omni-R2V Dataset}, a large-scale public R2V training dataset comprising 340K processed samples spanning heterogeneous reference types and multi-reference compositions. Built primarily from professional video footage through task-specific pipelines, it provides both an industry-grade training resource and reusable data construction pipelines for omni-R2V generation.
    \item We conduct an extensive evaluation of advanced open- and closed-source R2V models, revealing clear performance differences across tasks and evaluation dimensions and highlighting OmniVBench’s fine-grained evaluation capability.
\end{itemize}

\section{Related Work}
\noindent
\textbf{Benchmarks for Reference-to-Video Generation.}
Recent video generation is moving beyond text-only synthesis~\citep{cogvideox,Ltx-video,wan,longcatvideo,Hunyuanvideo_1.5,Waver_Arxiv25,veo31,pixelwizard} toward more flexible reference-conditioned generation.
Accompanying this shift, both commercial systems ~\citep{sora2,Vidu,skyreelsv3,seedance2_0,seedance2_5,kling_omni,gemini_omni,happy_horse,minimax_h3} and research models~\citep{VideoAlchemist,VACE,hunyuancustom,univideo,Stand-In,HuMo,Lynx,omnishow,OmniVCus,OmniWeaving_IntelligentVBench,bernini,loomvideo,Vino,Dreamid-omni_bytedance,PoCo,harmoview} have developed rapidly, supporting an increasingly diverse range of reference inputs and generation tasks.
This progress has also motivated new benchmarks for R2V generation~\citep{OmniWeaving_IntelligentVBench,UniVBench,OpenS2V-Nexus,MultiRef-Compass}.
However, these existing R2V benchmarks primarily focus on content references and their composition, leaving many reference conditions common in real-world creative workflows—such as motion, camera, style, layout, story, and continuation—largely underexplored. Their test cases and evaluation protocols also focus mainly on overall reference similarity and prompt alignment, with limited assessment of factor-level reference understanding and control.

\begin{figure*}[t]
\centering{\includegraphics[width=0.99\textwidth]{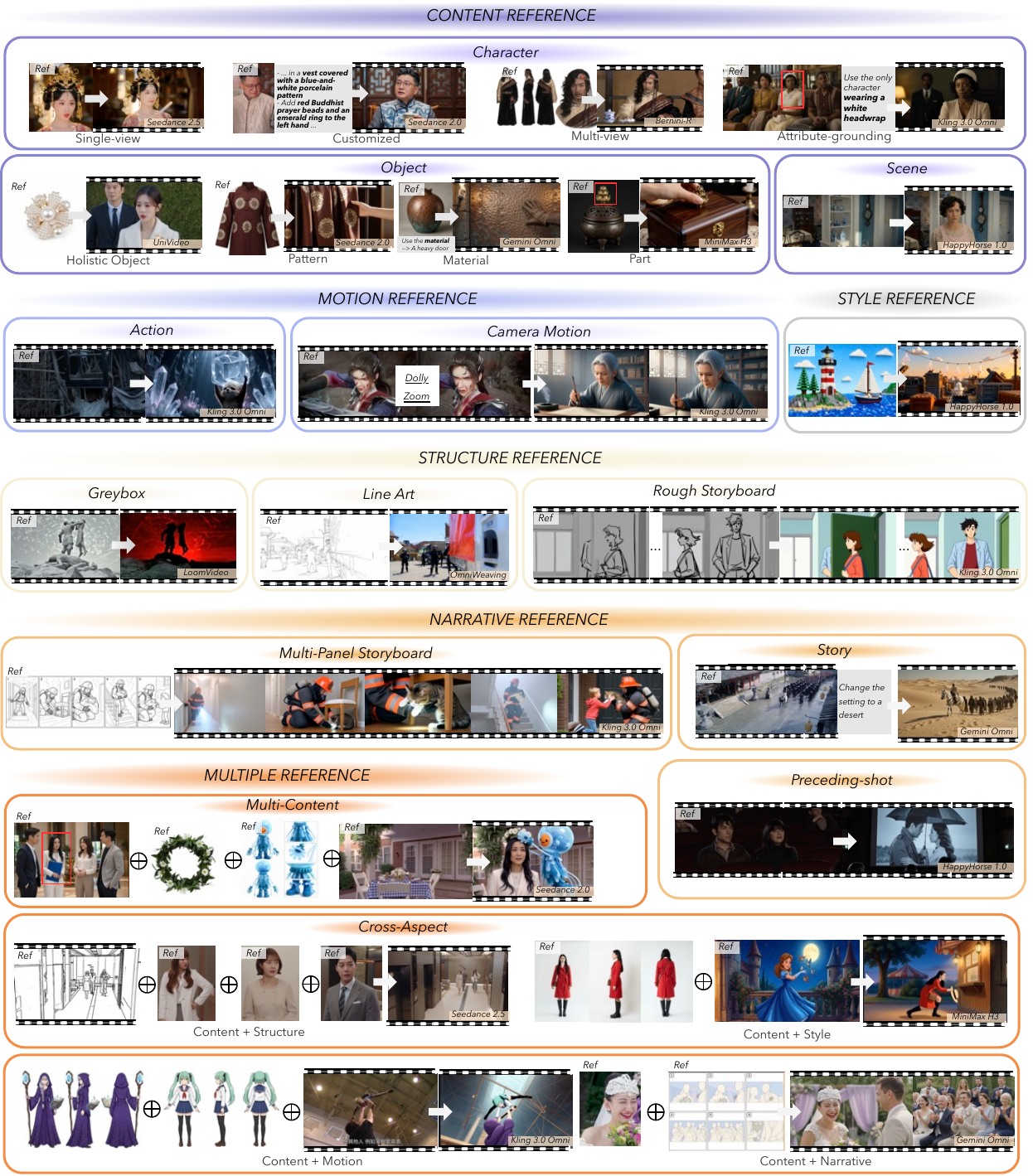}}
\caption{Representative examples of OmniVBench. Each category encompasses fine-grained generation tasks defined by distinct reference types and control objectives, with examples illustrating the reference inputs and corresponding generated outputs.}
\label{Fig.task}
\end{figure*}

\noindent
\textbf{Training Data for Reference-to-Video Generation.}
Existing large-scale video generation datasets are predominantly designed for text-to-video~\cite{OpenVid-1M,Koala-36M,OpenHumanVid, MiraData} and video editing~\cite{coinve200k,Openve-3m,senorita-2m,Ditto-1M}, whereas large-scale training resources for general R2V generation remain limited.
Some efforts have constructed datasets for reference-conditioned generation~\citep{OpenS2V-Nexus,phantom-data,OmniVCus}
However, these datasets remain largely centered on content references, with limited coverage of diverse reference types and compositional settings.
To address this gap, we introduce a large-scale dataset spanning heterogeneous reference factors, diverse instruction operations, and both single- and multi-reference settings, supporting the training of R2V models for a broader range of real-world creative workflows.

\section{OmniVBench}

OmniVBench is designed along two complementary axes: broad task coverage and factor-grounded evaluation. We first construct a reference-centric taxonomy covering diverse reference types, control settings, and their compositions (Sec.~\ref{sec:3.1}), and instantiate it into factor-aware evaluation cases that isolate the intended role of each reference (Sec.~\ref{sec:3.2}). We then evaluate model outputs with case-specific checklists grounded in these reference roles and the corresponding instructions (Sec.~\ref{sec:3.3}).

\subsection{Reference-Centric Task Taxonomy}
\label{sec:3.1}
We organize OmniVBench into five categories under the single-reference setting—content, motion, style, structure, and narrative—and further extend the taxonomy to multiple-reference settings, including multi-content and cross-aspect references (Fig.~\ref{Fig.task}).

\subsubsection{Single-Reference Tasks}

\paragraph{Content Reference.}
Content tasks evaluate how models use referenced entities and environments under different control settings. The referenced visual information is not explicitly specified in the textual instruction.
\textit{Object reference} is decomposed into holistic object identity, material, pattern, and part-level reference. This separation distinguishes coarse semantic copying from localized attribute transfer. \textit{Character reference} covers four settings. Single-view and multi-view reference test identity extraction under different visual coverage. 
Customized character reference provides an original character image together with an instruction that edits specified attributes—such as clothing—and requires the model to generate the target video with the modified character while preserving the remaining identity cues.  
Attribute-grounded character reference instead provides a multi-character scene and identifies the target through a spatial or visual description, requiring the model to first ground the correct person and then preserve that person’s identity in the generated video. 
\textit{Scene reference} evaluates environmental identity and layout while allowing instructed changes to foreground content.

\paragraph{Motion Reference.}
Motion tasks use video references and require models to transfer temporal dynamics while changing the original appearance and context. \textit{Action reference} tests whether subject motion can be disentangled from identity, background, and camera movement and then applied to a prompt-specified subject. The benchmark spans motions with varying spatial extent, temporal precision, and dynamic complexity, from routine actions to subtle articulations and highly dynamic performances. 
\textit{Camera-motion reference} instead transfers only the viewpoint trajectory, covering primitives such as pan, tilt, dolly, truck, pedestal, and orbit, as well as whip-pans, dolly-zooms, and compound movements. 
As the referenced motion is not explicitly described in the instruction, these tasks require models to extract and selectively transfer the intended motion from the reference video.

\paragraph{Style Reference.}
Style tasks pair a single reference image with an instruction that specifies the target content while leaving the reference style undescribed, assessing whether models can disentangle visual style from semantic content and transfer the intended style to the requested content. The references span diverse artistic media and visual traditions, including hand-drawn, painterly, animation, craft, digital, graphic, and cinematic styles.

\paragraph{Structure Reference.} 
Structure tasks provide temporally ordered guidance with incomplete appearance details. We use \textit{Greybox}, \textit{Line Art}, and \textit{Rough Storyboards} as reference forms, and request 2D animation, 3D animation, or live-action outputs where applicable. 
Models are required to preserve the referenced spatial and temporal structure while completing missing appearance details according to the instruction.

\paragraph{Narrative Reference.}
Narrative tasks evaluate whether a model can extract story structure from a reference and follow, transform, or continue it as instructed. 
\textit{Multi-panel storyboard reference} provides an ordered storyboard grid and requires the model to infer its narrative progression and realize it as a continuous video. 
\textit{Story reference} uses an existing video as a narrative reference, requiring the model to follow its storyline while allowing changes to characters, settings, and visual appearance.
\textit{Preceding-shot reference} provides the preceding shot as context and asks the model to generate a coherent next shot according to a textual brief. 

\subsubsection{Multi-Reference Tasks}

We consider two settings: multi-content, where multiple references specify different entities or attributes, and cross-aspect, where references control complementary aspects of the output.

\paragraph{Multi-content Reference.}
This setting includes two configurations. \textit{Direct entity composition} uses separate references for characters, objects, and scenes, and requires all specified entities to appear in their assigned roles. \textit{Grounded factor composition} further introduces references containing multiple candidate entities or attributes. The model must locate the instructed target, selectively transfer factors, and bind each factor to the correct output component.

\noindent
\textbf{Cross-aspect Reference.}
This setting combines references that control different aspects of the output, including content with motion, style, structure, or narrative. The model must extract the designated factor from each source and satisfy the conditions jointly without allowing one reference to overwrite another. For example, content--motion tasks pair appearance images with a motion video depicting a different subject, while content--structure tasks combine entity references with line art or storyboards.

\begin{figure*}[t]
\centering{\includegraphics[width=0.99\textwidth]{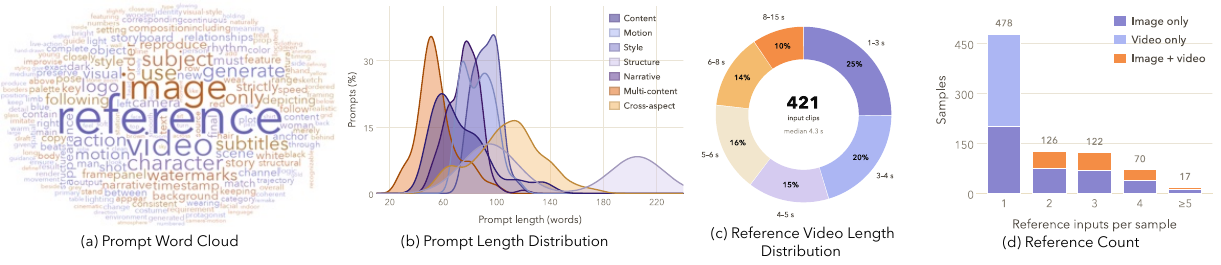}}
\caption{Statistics of OmniVBench. We present the distribution of benchmark samples from multiple perspectives, including (a) the word cloud of textual prompts, (b) prompt length distributions across task categories, (c) reference video length distribution, and (d) the number and modality composition of reference inputs per sample.}
\label{Fig.statistic_bmk}
\vspace{-10pt}
\end{figure*}
\subsection{Benchmark Construction and Statistics}
\label{sec:3.2}

\paragraph{Case Design.}
Given the taxonomy above, we construct each evaluation case so that the intended reference information must be inferred from the reference rather than duplicated in the textual instruction. Depending on the task, the instruction may instead specify how that information should be preserved, modified, grounded, or composed in the target video.

\paragraph{Data Construction.}
Videos are sourced from a large-scale internal collection with the necessary rights and permissions for research use and public release. 
References are obtained through cross-segment matching or task-specific construction.
All samples undergo manual verification for reference quality and task consistency. Detailed task definitions and construction procedures are provided in Appendix~\ref{app:benchmark_construction_details}.

\paragraph{Benchmark Statistics.}
The statistics of OmniVBench are summarized in Fig.~\ref{fig:teaser-top} and Fig.~\ref{Fig.statistic_bmk}.
OmniVBench comprises 813 evaluation cases spanning 18 sub-tasks across diverse reference types and compositional settings. 
The cases cover diverse instruction lengths and reference-video durations, ranging from single image/video references to multiple image–video combinations.

\subsection{Evaluation Protocol}
\label{sec:3.3}

\subsubsection{Evaluation Capability Taxonomy}
Given these factor-aware cases, our evaluation explicitly assesses whether each intended reference factor is correctly used according to its designated role. We decompose evaluation into three dimensions: Reference Fidelity (RF), Instruction Realization (IR), and Video Quality (VQ).

\noindent
\textbf{Reference Fidelity (RF)} measures how faithfully the designated reference information is preserved or transferred to the generated video. It comprises five L2 sub-dimensions: \textit{content}, \textit{structure}, \textit{motion}, \textit{style}, and \textit{narrative} fidelity. For each case, only the relevant sub-dimensions are evaluated, with factor-grounded checklists assessing the specific reference factors required by the case. Changes explicitly specified by the instruction are not considered fidelity errors.

\noindent
\textbf{Instruction Realization (IR)} measures whether the requested operations are correctly realized and comprises two L2 sub-dimensions. 
\textit{(1) Reference-Factor Disentanglement and Routing} evaluates whether the intended reference information is selectively transferred without copying irrelevant factors and assigned to the correct target.
\textit{(2) Target Compliance} evaluates whether the additional requirements specified by the instruction are correctly realized.
For each sample, only the relevant IR sub-dimensions are evaluated using sample-specific questions.

\noindent
\textbf{Video Quality (VQ)} evaluates the output independently of reference fidelity and instruction compliance. It comprises three L2 sub-dimensions: \textit{Technical Quality}, \textit{Aesthetic Quality}, and \textit{Physical Plausibility}. We evaluate these dimensions using the Technical branch of DOVER++~\cite{dover}, Aesthetic Predictor V2.5~\cite{Aes_predictor_v2.5}, and the Coherence/Physics dimension of UnifiedReward 2.0~\cite{unifiedreward}, respectively. 

\subsubsection{Factor-Grounded Checklist Evaluation}
For RF and IR, each L2 sub-dimension is associated with a fixed set of evaluation criteria (Appendix~\ref{app:l2_criteria}). 
Given a sample, the relevant criteria are first selected by a VLM-based criterion activator~\citep{gemini_3_1_pro} and instantiated as atomic checklists based on the instruction, references, reference roles, and target bindings. 
Multiple reference factors are evaluated separately, and all checklists are manually verified before evaluation.
A VLM~\citep{gemini_3_1_pro} then evaluates RF and IR in separate calls using the instruction, references, generated video, and the corresponding checklist and scoring rubric.
RF questions are rated on a 1–5 scale, ranging from no meaningful correspondence to complete and temporally consistent reproduction, while IR questions are rated as failed (1), partially realized (2), or fully realized (3).

\paragraph{Score Aggregation.}
For RF and IR, scores from checklists belonging to the same criterion are first averaged and linearly mapped to a 100-point scale. 
For a sample $i$ and dimension $d\in\{\mathrm{RF},\mathrm{IR}\}$, let $G_i^d$ denote its active L2 sub-dimensions, $A_{i,g}$ the active criteria under sub-dimension $g$, and $z_{i,c}$ the normalized score for criterion $c$. We compute $
S_i^d =
\frac{1}{|G_i^d|}
\sum_{g\in G_i^d}
\frac{1}{|A_{i,g}|}
\sum_{c\in A_{i,g}} z_{i,c}.
$
Within each task family, sample-level RF, IR, and VQ scores are averaged separately. The family score averages these three dimensions equally. 

\begin{figure*}[t]
\centering{\includegraphics[width=0.9\textwidth]{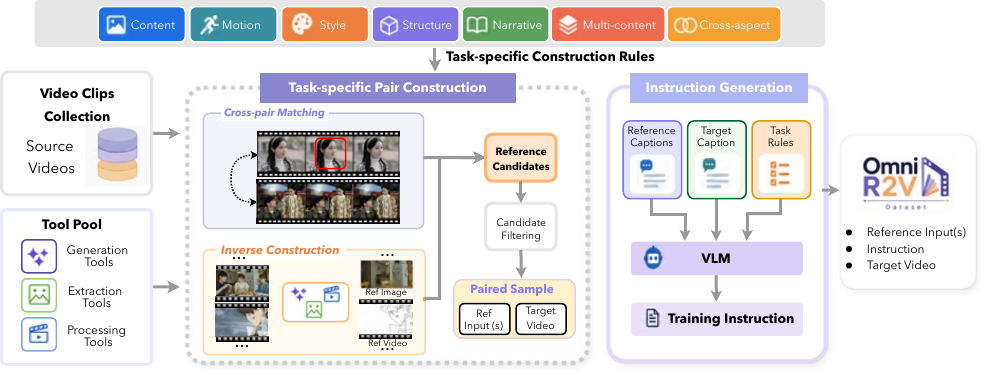}}
\vspace{-5pt}
\caption{Overview of the Omni-R2V Dataset construction pipeline. }
\label{Fig.data_pipe}
\end{figure*}

\section{Omni-R2V Dataset}

We construct Omni-R2V, a large-scale, high-quality dataset built from a professionally curated video corpus and spanning the seven task families in OmniVBench. 
The central challenge is to establish task-specific reference–target relationships across heterogeneous reference types and express these relationships through corresponding training instructions. 
As illustrated in Fig.~\ref{Fig.data_pipe}, our pipeline consists of source-video processing, task-specific pair construction, candidate filtering, and instruction generation.

\subsection{Dataset Statistics}
As summarized in Fig.~\ref{fig:teaser-top}, Omni-R2V dataset contains 339,570 training instances (approximately 340K) across seven task families, covering 2D animation, 3D animation, and live-action videos. Clip durations extend to 20 seconds, with distributions varying across task families. Resolutions range from \(480\mathrm{p}\) to \(2160\mathrm{p}+\), with 1080p and 2160p+ accounting for the majority.

\subsection{Data Construction Pipeline}
\paragraph{Source Video Collection.} 
Omni-R2V is built primarily from an in-house video collection, supplemented with publicly available data~\cite{Koala-36M}. We segment source videos into temporally coherent clips and filter them for visual quality, event completeness, and suitability for R2V training. The retained clips provide both target videos and source material for constructing heterogeneous image and video references.

\paragraph{Task-specific pair construction.}
For each target video, task-specific rules determine the required reference roles and reference--target relationships. 
We construct reference candidates through two complementary strategies:
\emph{cross-pair matching} and \emph{inverse construction}.
Cross-pair matching selects reference clips or frames from existing videos
according to the relationship required by the task, such as shared character
identity or stylistic consistency.
Inverse construction derives references from the target through extraction~\cite{canny,ComfyUI-Anyline,hed,pidinet,lineart_realistic} or generation~\cite{qwen_image_edit,gpt_image_2,VACE,Wan-animate}, preserving task-relevant information while modifying or simplifying other aspects where appropriate. For multi-reference tasks, we combine references obtained through either strategy, with each reference specifying a designated entity or visual factor in the same target video. Reference candidates undergo task-specific filtering to verify that the required information is preserved and visually identifiable before being assembled into reference--target pairs. Detailed task-specific procedures are provided in Appendix~\ref{sec:appendix_omni_r2v_construction}.

\paragraph{Instruction Generation.} 
After fixing the references and their roles, Gemini-3.1-Pro~\cite{gemini_3_1_pro} captions each reference and the target video separately. DeepSeek-V4-Pro~\cite{DeepSeek-V4-Pro} then converts these descriptions into a training instruction using task-specific rules. The instruction specifies the requested target content, identifies each reference and its intended role, and makes the correspondence between references and target entities or factors explicit.

\section{Experiment}

\subsection{Experimental Setup}

We evaluate a diverse set of state-of-the-art Omni-R2V models, including closed-source models:
Kling 3.0 Omni~\citep{kling_omni}, Seedance 2.0~\citep{seedance2_0}, Seedance 2.5~\citep{seedance2_5}, Gemini Omni~\citep{gemini_omni}, Happy Horse 1.0~\citep{happy_horse}, and Vidu-Q2-Pro~\citep{Vidu}, as well as the open-source models: 
UniVideo~\citep{univideo}, OmniWeaving~\citep{OmniWeaving_IntelligentVBench}, LoomVideo~\citep{loomvideo}, Bernini~\citep{bernini},  and MiniMax H3~\citep{minimax_h3}.
For MiniMax H3, we use the official H3-Context-IR to preprocess the multimodal references and instructions before generation, following the recommended workflow.
%API-rejected cases are treated as unavailable outputs rather than generation failures. Accordingly, the average score is computed over valid model responses only, so that differences in safety filtering do not confound the comparison of generative performance.

\definecolor{OursOrange}{HTML}{F4BB6E}
\definecolor{SubHeaderBlue}{RGB}{240,246,252}
\definecolor{FirstPlace}{HTML}{ACB6F3}

\begin{table*}[!t]
\centering
\caption{
Fine-grained comparison of different models on OmniVBench.
Each task score is averaged over Reference Fidelity (RF),
Instruction Realization (IR), and Visual Quality (VQ). The \textbf{first-place} and \underline{second-place} results are highlighted accordingly.
}
\label{tab:main_results}

\setlength{\tabcolsep}{18pt}
\resizebox{\textwidth}{!}{%
\begin{tabular}{l*{8}{c}}
\toprule

\textbf{Model}
& \textbf{Content}
& \textbf{Motion}
& \textbf{Style}
& \textbf{Structure}
& \textbf{Narrative}
& \makecell[c]{\textbf{Multi-}\\\textbf{Content}}
& \makecell[c]{\textbf{Cross-}\\\textbf{Aspect}}
& \textbf{Overall}
\\

\midrule

\multicolumn{9}{c}{\textbf{\textit{Open-Source Models}}} \\
\midrule

UniVideo~\citep{univideo}
& 66.01 & 44.52 & 47.40 & 48.84 & 36.97 & 58.00 & 46.09 & 49.69 \\

OmniWeaving~\citep{OmniWeaving_IntelligentVBench}
& 62.23 & 50.58 & 41.40 & 63.19 & 38.03 & 47.49 & 39.25 & 48.88 \\

LoomVideo~\citep{loomvideo}
& 61.24 & 52.09 & 42.03 & 62.53 & 46.98 & 49.02 & 44.86 & 51.25 \\

Bernini~\citep{bernini}
& 69.27 & 55.92 & 50.62 & 69.44 & 42.02 & 60.32 & 51.03 & 56.95 \\

MiniMax H3~\citep{minimax_h3}
& 78.86
& 65.32
& 66.24
& \underline{72.94}
& 73.90
& \underline{77.22}
& \textbf{72.36} & \underline{72.41} \\

\midrule

\multicolumn{9}{c}{\textbf{\textit{Closed-Source Models}}} \\
\midrule

Vidu-Q2-Pro~\citep{Vidu}
& 71.87 & 47.65 & 59.87 & 51.81 & 50.68 & 72.07 & 62.21 & 59.45 \\

Kling 3.0 Omni~\citep{kling_omni}
& 75.99 & 64.20 & 55.90 & 71.12 & 68.59 & 75.98 & 68.33 & 68.59 \\

Happy Horse 1.0~\citep{happy_horse}
& 75.90
& \textbf{68.58}
& 65.62
& 69.46
& 72.34
& 74.62
& 69.71 & 70.89 \\

Gemini Omni~\citep{gemini_omni}
& 75.02
& 62.25
& \textbf{69.24}
& 70.13
& \textbf{75.11}
& 73.36
& 70.18 & 70.76 \\

Seedance 2.0~\citep{seedance2_0}
& \textbf{79.00}
& 63.05
& 63.73
& 68.36
& 73.76
& 76.63
& \underline{71.83} & 70.91 \\

Seedance 2.5~\citep{seedance2_5}
& \underline{78.88}
& \underline{66.10}
& \underline{67.13}
& \textbf{73.35}
& \underline{73.97}
& \textbf{77.77}
& 71.53 & \textbf{72.68} \\

\bottomrule
\end{tabular}%
}
\vspace{-6pt}
\end{table*}

\subsection{Results}

\noindent
\textbf{Overall Performance on OmniVBench.}
Table~\ref{tab:main_results} presents the overall performance of representative open- and closed-source models.
The strongest open-source models now achieve performance comparable to several leading closed-source systems.
Current models generally achieve higher scores on content-reference tasks, while performance is lower or more variable across motion, style, structure, narrative, and multi-reference settings.
This variation suggests that progress on one reference condition does not necessarily transfer to others, highlighting the diverse capability requirements of omni R2V generation.

\begin{wrapfigure}{r}{0.6\columnwidth}
\centering
\vspace{-6pt}
\includegraphics[width=0.59\columnwidth]{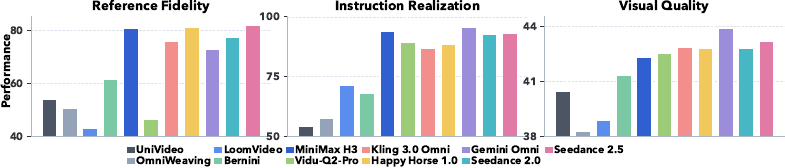}
\caption{Model performance across RF, IR, and VQ.}
\label{fig:rf_ir_vq}
\vspace{-12pt}
\end{wrapfigure}

\paragraph{Fine-Grained Capability Analysis.}
Fig.~\ref{fig:rf_ir_vq} further compares model performance along the three evaluation dimensions: Reference Fidelity (RF), Instruction Realization (IR), and Video Quality (VQ). 
Seedance 2.5 achieves particularly strong RF, whereas models such as Vidu-Q2-Pro remain competitive in VQ despite less prominent performance in reference control. Among open-source models, MiniMax H3 consistently outperforms the other models across all three dimensions, while the remaining open-source models show substantial room for improvement in RF, IR, and VQ.

\begin{figure*}[t]
\centering{\includegraphics[width=0.9\textwidth]{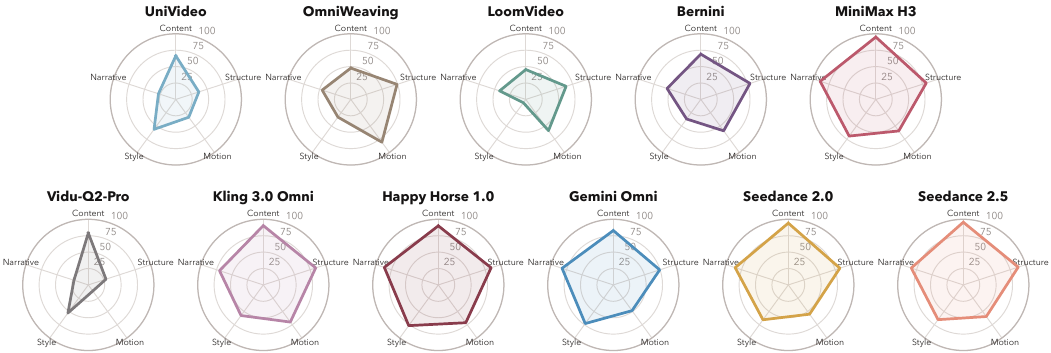}}
\caption{Fine-grained comparison of Reference Fidelity (RF) across RF sub-dimensions. }
\label{Fig.compare_RF_radar}
\end{figure*}

\paragraph{Reference Fidelity Analysis.} Fig.~\ref{Fig.compare_RF_radar} further breaks down RF into its five L2 sub-dimensions: content, structure, motion, style, and narrative fidelity. 
Across models, content fidelity is generally stronger, while larger performance gaps emerge for motion, structure, and narrative fidelity. 
This gap indicates that current models are considerably better at preserving visual content than more complex reference information involving temporal dynamics, structural constraints, and narrative progression, highlighting the need to assess these reference factors separately.

% Required packages:
% \usepackage{booktabs}
% \usepackage{multirow}
% \usepackage{graphicx}
% \usepackage[table]{xcolor}

\definecolor{OursOrange}{HTML}{F4BB6E}
\definecolor{FirstPlace}{HTML}{ACB6F3}
\definecolor{SubHeaderBlue}{RGB}{240,246,252}

\begin{table*}[t]
\centering
\caption{
Fine-grained comparison of Instruction Realization (IR) on OmniVBench. $\mathrm{IR}_{\mathrm{DR}}$ measures reference-factor disentanglement and routing, while $\mathrm{IR}_{\mathrm{TC}}$ measures target compliance. 
}
\label{tab:ir_fine_grained}

\setlength{\tabcolsep}{12pt}
\renewcommand{\arraystretch}{1.12}

\resizebox{\textwidth}{!}{%
\begin{tabular}{l*{14}{c}}
\toprule
\multirow{2}{*}{\textbf{Model}}
& \multicolumn{2}{c}{\textbf{Content}}
& \multicolumn{2}{c}{\textbf{Motion}}
& \multicolumn{2}{c}{\textbf{Style}}
& \multicolumn{2}{c}{\textbf{Structure}}
& \multicolumn{2}{c}{\textbf{Narrative}}
& \multicolumn{2}{c}{\textbf{Multi-Content}}
& \multicolumn{2}{c}{\textbf{Cross-Aspect}} \\
\cmidrule(lr){2-3}
\cmidrule(lr){4-5}
\cmidrule(lr){6-7}
\cmidrule(lr){8-9}
\cmidrule(lr){10-11}
\cmidrule(lr){12-13}
\cmidrule(lr){14-15}
& $\mathrm{IR}_{\mathrm{DR}}$ & $\mathrm{IR}_{\mathrm{TC}}$
& $\mathrm{IR}_{\mathrm{DR}}$ & $\mathrm{IR}_{\mathrm{TC}}$
& $\mathrm{IR}_{\mathrm{DR}}$ & $\mathrm{IR}_{\mathrm{TC}}$
& $\mathrm{IR}_{\mathrm{DR}}$ & $\mathrm{IR}_{\mathrm{TC}}$
& $\mathrm{IR}_{\mathrm{DR}}$ & $\mathrm{IR}_{\mathrm{TC}}$
& $\mathrm{IR}_{\mathrm{DR}}$ & $\mathrm{IR}_{\mathrm{TC}}$
& $\mathrm{IR}_{\mathrm{DR}}$ & $\mathrm{IR}_{\mathrm{TC}}$ \\
\midrule

UniVideo~\cite{univideo}
& 62.32 & 73.47
& 57.27 & 70.14
& 15.39 & 50.78
& 71.13 & 82.80
& 26.03 & 33.62
& 57.48 & 54.68
& 53.23 & 53.92 \\

OmniWeaving~\cite{OmniWeaving_IntelligentVBench}
& 78.36 & 85.87
& 39.95 & 51.05
& 59.33 & 67.13
& 67.23 & 92.09
& 23.78 & 36.35
& 48.55 & 67.66
& 40.55 & 54.20 \\

LoomVideo~\cite{loomvideo}
& 87.91 & 88.65
& 57.28 & 68.38
& 89.64 & 68.42
& 76.62 & 93.95
& 58.88 & 37.48
& 65.00 & 74.09
& 55.83 & 63.27 \\

Bernini~\cite{bernini}
& 83.19 & 91.90
& 72.65 & 61.46
& 68.15 & 72.58
& 77.54 & 97.11
& 28.57 & 37.81
& 65.77 & 78.01
& 55.94 & 64.08 \\

MiniMax H3~\cite{minimax_h3}
& 98.21 & \textbf{97.28}
& 94.14 & 94.99
& 91.94 & 85.65
& 95.42 & 98.14
& 95.05 & \textbf{89.47}
& \textbf{95.93} & 97.80
& \textbf{93.61} & \textbf{90.71} \\

Vidu-Q2-Pro~\cite{Vidu}
& 95.47 & 93.07
& 86.85 & 89.81
& 96.55 & 77.67
& 91.98 & 96.92
& 84.69 & 56.59
& 91.46 & 92.46
& 86.77 & 77.77 \\

Kling 3.0 Omni~\cite{kling_omni}
& 91.87 & 94.16
& 84.37 & 82.62
& 78.51 & 77.51
& 92.67 & 91.45
& 87.30 & 65.91
& 91.44 & 92.07
& 86.10 & 81.96 \\

Happy Horse 1.0~\cite{happy_horse}
& 93.68 & 95.20
& 80.74 & 92.30
& 84.26 & \textbf{89.07}
& 75.13 & 97.15
& 83.86 & 81.50
& 94.78 & 95.57
& 85.71 & 83.52 \\

Gemini Omni~\cite{gemini_omni}
& \textbf{98.32} & 97.10
& \textbf{94.82} & \textbf{96.07}
& 95.29 & 90.71
& \textbf{96.66} & \textbf{98.64}
& \textbf{96.96} & 85.26
& 95.19 & \textbf{98.45}
& 91.67 & 90.13 \\

Seedance 2.0~\cite{seedance2_0}
& 96.85 & 95.44
& 91.49 & 95.88
& \textbf{100.00} & 83.89
& 81.31 & 94.59
& 90.38 & 85.52
& 94.29 & 96.68
& 91.83 & 89.59 \\

Seedance 2.5~\cite{seedance2_5}
& 96.77 & 96.71
& 93.60 & 94.84
& 96.09 & 85.23
& 87.90 & 96.95
& 89.04 & 72.66
& 94.52 & 95.40
& 92.39 & 87.83 \\

\bottomrule
\end{tabular}%
}
\end{table*}

\begin{figure}[!ht]
\centering
% ===== Left ====
\begin{minipage}[t]{0.49\linewidth}
    \centering
    \setlength{\tabcolsep}{3mm}{
    \captionof{table}{Correlation between automatic evaluation and human judgments.
    }
    \label{tab:evaluation_validation}
    \scalebox{0.6}{
    \begin{tabular}{lcc}
    \toprule
    \textbf{Dimension} & \textbf{Pearson} & \textbf{Spearman} \\
    \midrule
    Reference Fidelity (RF)     & 0.81 & 0.77 \\
    Instruction Realization (IR) & 0.78 & 0.74 \\
    Video Quality (VQ)           & 0.86 & 0.82 \\
    \bottomrule
    \end{tabular}}}
\end{minipage}
\hfill
% ===== Right =====
\begin{minipage}[t]{0.48\linewidth}
    \centering
    \setlength{\tabcolsep}{3mm}{
    \captionof{table}{
    Ablation of factor-grounded checklist evaluation.
    }
    \label{tab:factor_grounded_checklist}
    \scalebox{0.6}{
    \begin{tabular}{lcc}
    \toprule
    \textbf{Eval. Method} & \textbf{RF Spearman} & \textbf{IR Spearman} \\
    \midrule 
    Holistic & 0.67 & 0.69  \\
    Factor-grounded checklist & 0.77 & 0.74 \\
    \bottomrule
  \end{tabular}}}
\end{minipage}
\end{figure}
\vspace{-3pt}

\paragraph{Instruction Realization Analysis.}
Table 4 further decomposes IR into its two L2 sub-dimensions, reference-factor disentanglement and routing ($\mathrm{IR}_{\mathrm{DR}}$) and target compliance ($\mathrm{IR}_{\mathrm{TC}}$). 
The results reveal clear gaps between the two capabilities across models and task families. The two capabilities show different patterns across tasks: Happy Horse 1.0 shows stronger target compliance than disentanglement and routing on structure reference, whereas Vidu-Q2-Pro shows the opposite trend on narrative reference. This indicates that strong performance in one capability does not necessarily lead to strong performance in the other, supporting the need to evaluate them separately.

\paragraph{Evaluation Protocol Validation.}
To validate our evaluation protocol, we sample 100 benchmark cases covering all seven task families and 18 sub-tasks. Three human annotators independently rate each output on Reference Fidelity (RF), Instruction Realization (IR), and Video Quality (VQ) using a five-point scale.
As shown in Table~\ref{tab:evaluation_validation}, our automatic evaluation shows strong agreement with human judgments across RF, IR, and VQ. Furthermore, Table~\ref{tab:factor_grounded_checklist} shows that the factor-grounded checklist achieves consistently higher human correlation than holistic evaluation, demonstrating the benefit of explicitly assessing case-specific reference factors and instruction requirements.

\section{Conclusion}
We present \textbf{OmniVBench}, a comprehensive benchmark for omni reference-to-video generation, together with the \textbf{Omni-R2V Dataset}, a large-scale collection of 340K processed R2V training samples spanning diverse reference types and their compositions. OmniVBench combines broad task coverage with factor-grounded evaluation to assess reference fidelity, instruction realization, and video quality, enabling diagnostic analysis of fine-grained reference control and multi-reference composition. Extensive evaluation of representative open- and closed-source R2V models provides insights into the strengths and limitations of current R2V models across diverse reference conditions. We hope OmniVBench and the Omni-R2V Dataset provide complementary evaluation and training resources for advancing more general and controllable R2V generation.

\clearpage

\bibliographystyle{colm2024_conference}
\bibliography{omnivbench}

\appendix
\clearpage
\section{Appendix}

\renewcommand{\thefigure}{\thesection.\arabic{figure}}
\renewcommand{\thetable}{\thesection.\arabic{table}}
\renewcommand{\theequation}{\thesection.\arabic{equation}}
\setcounter{figure}{0}
\setcounter{table}{0}
\setcounter{equation}{0}

\subsection{More Omni-R2V Dataset Details}

\subsubsection{Detailed Omni-R2V Dataset Construction}
\label{sec:appendix_omni_r2v_construction}

We provide the detailed construction procedures for each task family in the Omni-R2V Dataset.
Following the general pipeline described in Sec.4, reference--target pairs are constructed through cross-pair matching or inverse construction, with task-specific operations determined by the reference factors required for each task. The Omni-R2V dataset and the OmniVBench benchmark are non-overlapping.

\paragraph{Content reference.}
Content-reference pairs are constructed through cross-video identity matching. Given a target clip, we retrieve other clips or segments depicting the same character and sample reference frames in which the character is clearly visible. The reference and target are selected from different temporal segments whenever possible, so that they preserve identity while varying in pose, viewpoint, action, attire, styling, and surrounding content.

\paragraph{Motion reference.}
Action reference and camera motion reference data are constructed through separate pipelines. To construct action references, we first retrieve a compatible character image for each target video from a large-scale image pool based on coarse spatial attributes, including the character's position, orientation, and body scale within the frame. These constraints ensure that the retrieved character is spatially compatible with the target while differing in appearance and identity. We then use Wan-Animate~\citep{Wan-animate} to animate the retrieved character with the motion extracted from the target video, producing an action reference that preserves the target motion while varying its character content. 
We construct camera motion references from three complementary sources. First, we synthesize videos with controlled camera trajectories in Unreal Engine, providing references with explicitly specified camera motion. Second, we mine references from real videos by estimating their camera trajectories using an off-the-shelf camera pose estimation model~\citep{pi3}, encoding the resulting pose sequences into trajectory embeddings, and retrieving videos with similar camera-motion patterns. Third, we extract camera trajectories from target videos and use an internal camera-controllable video generation model to synthesize references that follow the target camera motion while varying the visual content.

\paragraph{Style reference.}
We construct style references through cross-pair matching. For each target video, we sample a frame from a different video pair originating from the same source video and retain it only if its event, subject, and environment are semantically unrelated to those of the target. This reduces content overlap between the reference and target while preserving stylistic consistency, allowing the reference to primarily provide stylistic rather than semantic cues.

\paragraph{Structure reference.}
For line art reference, we uniformly sample one of 12 frame-wise extractors from four families: Canny gradient edges~\cite{canny}, learned soft edges based on HED~\cite{hed}, PiDiNet~\cite{pidinet}, and TEED~\cite{TEED}, coarse scribbles produced by applying stronger binarization and thinning to the same HED or PiDiNet, and artist-oriented line drawings extracted with Informative Drawings~\cite{lineart_realistic} or AnyLine~\cite{ComfyUI-Anyline}. The same extractor is applied to all frames of a clip. For approximately half of the samples, we invert the line-map colors to include both dark lines on a light background and light lines on a dark background. We discard near-empty or excessively dense line maps that are difficult to interpret.
We construct greybox references through a two-stage pipeline. Given a target video, we first edit its initial frame into a greybox representation while preserving the scene geometry and spatial layout. We then use the edited first frame together with the depth sequence extracted from the target video as structural conditions for Wan2.2-VACE-Fun-A14B~\citep{wan22vacefun}, producing a temporally coherent greybox video that follows the structure and motion of the target.
For rough-storyboard reference, we first extract temporally ordered keyframes from each source video. Qwen-Image-Edit~\cite{qwen_image_edit} converts each keyframe into a sketch using the instruction ``Transform the image into a sketch''. The converted frames are then assembled in their original temporal order. Gemini-3.1-Pro~\cite{gemini_3_1_pro} verifies character identity, number, and position, as well as shot type, action, pose, expression, scene content, and object content. It also checks style consistency and character consistency across frames, residual text, and visible generation artifacts.

\paragraph{Narrative reference.}
Preceding-shot references are constructed from temporally adjacent shots in source videos. We first use TransNetV2~\cite{transnet_v2} to detect shot boundaries and segment each video into consecutive shots, after which Gemini-3.1-Pro~\cite{gemini_3_1_pro} evaluates each adjacent shot pair in terms of scene consistency, subject consistency, motion continuity, spatial consistency, and narrative continuity. Pairs that exhibit sufficient cross-shot coherence are retained, with the earlier shot serving as the preceding-shot reference and the subsequent shot as the target video. 
For multi-panel storyboard references, we sample multiple frames from the target video at different timestamps and arrange them into a single grid, providing a compact representation of its temporal progression and key visual states.

\paragraph{Multi-Content Reference.}
We construct multi-content references by mixing references obtained through cross-segment matching and inverse construction, covering character–scene, character–object, and character–object–scene combinations. Character references are obtained through cross-segment matching. Object and scene references are extracted from source segments and evaluated by a VLM to ensure that the corresponding objects or scenes are clearly visible. We then use GPT Image 2.0~\cite{gpt_image_2} to perform inverse construction, producing isolated object images or background-only images. The resulting character, object, and scene references are finally combined to form the corresponding multi-content reference configurations.

\paragraph{Cross-Aspect Reference.}
We construct cross-aspect references by combining complementary reference construction strategies. For content–style references, we first construct the style reference following the style-reference pipeline and then perform inverse construction to generate a subject with the same identity but a different visual style. For content–structure references, we first construct the structure reference following the structure-reference pipeline and then obtain the content reference through cross-segment matching, ensuring that the two references provide complementary content and structural information.

\begin{figure*}[t]
\centering{\includegraphics[width=0.99\textwidth]{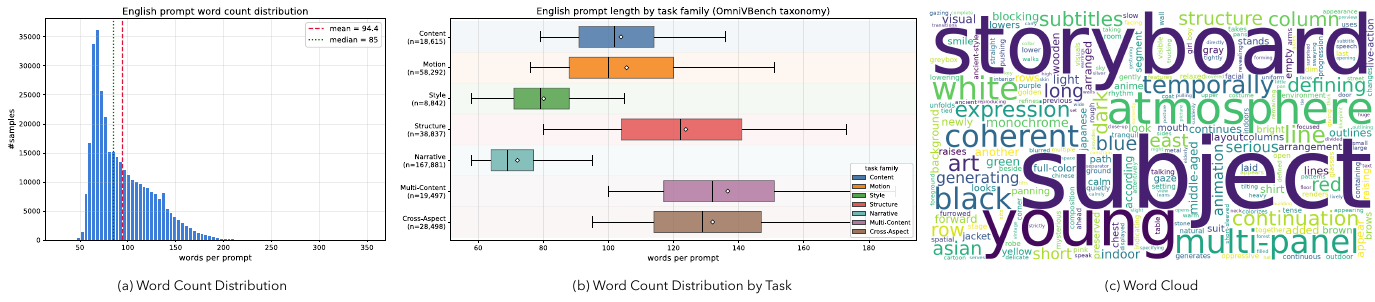}}
\caption{Prompt statistics of \textbf{Omni-R2V Dataset}.}
\label{Fig.prompt_statistics}
\end{figure*}
\subsubsection{Prompt Statistics}
Fig.~\ref{Fig.prompt_statistics} summarizes the English prompts from three perspectives. 
The overall word-count distribution in (a) shows that the dataset covers prompts with varying levels of descriptive and instructional detail. 
The task-wise distributions in (b) further reveal differences in prompt length across the seven task families, reflecting their different reference configurations and instruction requirements. 
The word cloud in (c) provides a qualitative view of the vocabulary covered by the dataset, including diverse subjects, scenes, attributes, actions, spatial concepts, and visual descriptions.

\subsection{OmniVBench Construction Details}
\label{app:benchmark_construction_details}
We detail the construction procedures for each task family in OmniVBench below, including reference selection, generation, and task-specific design. All benchmark samples are reviewed by three human annotators to verify visual quality, the clarity of the intended reference information, and consistency with the corresponding task requirements.

\paragraph{Content reference.}
Content references are constructed through cross-segment identity matching, following Sec.~\ref{sec:appendix_omni_r2v_construction}, or generated with GPT Image 2.0~\cite{gpt_image_2}.
The construction procedure varies by task:
\begin{itemize}
\item \textit{Object.} Object references are generated with GPT Image 2.0~\cite{gpt_image_2} according to the specific reference attributes required by each task. These cases test whether models can preserve whole-object identity or selectively transfer a specified material, pattern, or part, depending on the instruction.
\item \textit{Character.} For character reference, single-view and customized character samples are obtained by matching segments of the same character across different clips, with task-specific attribute modifications applied to customized characters. Multi-view references are generated with GPT Image 2.0 to provide complementary views of the same character. For attribute-grounded reference, we extract textual descriptions of the target character’s distinctive visual attributes and combine the target with 2–4 characters with distinguishable appearances to generate multi-character reference images. Together, these settings test identity preservation across viewpoints, selective attribute modification, and identification of the intended character among visually distinct distractors.
\item \textit{Scene.} Scene references are generated from source-video content using GPT Image 2.0. These cases test whether models can preserve the reference environment’s distinctive appearance and spatial layout while accommodating instructed changes to foreground content.
\end{itemize}
All samples are filtered to ensure that the intended reference information is visually identifiable and not explicitly revealed in the textual instruction.

\paragraph{Motion reference.}
We first define representative motion categories and then manually select source videos whose motion patterns match the corresponding definitions. Motion reference is divided into subject action and camera motion. 
\begin{itemize}
    \item \textit{Action.} The action-reference set covers locomotion, upper-body manipulation, daily activities, work-related actions, children's play, animal motion, fine-grained hand movement, facial expressions, dance, sports, combat, and acrobatics. This range tests whether models can transfer actions with different spatial scales and temporal complexity to a new subject while separating the action from the reference subject’s appearance and surroundings.
    \item \textit{Camera Motion.} We select reference videos covering atomic camera-motion primitives---including pan, tilt, dolly, truck, tracking, pedestal, orbit, and whip-pan---as well as static shots, dolly zooms, and complex multi-stage camera sequences. These cases test whether models can distinguish camera movement from subject motion and reproduce the referenced viewpoint trajectory, including transitions between successive camera movements.
\end{itemize}

\paragraph{Style Reference.}
We curate 43 representative visual styles across 16 categories. The collection covers traditional Eastern art (\eg, ink wash, gongbi, Chinese art animation, shadow puppetry, Dunhuang murals, and ukiyo-e), hand-drawn and painterly media (e.g., watercolor, pencil sketch, impasto, woodcut, and crayon), handcrafted and material-based art (e.g., paper collage, patchwork, clay, felt, and stained glass), Japanese animation and Western cartoon aesthetics, digital and graphic styles (e.g., pixel art, pop art, vaporwave, low-poly rendering, and geometric illustration), and cinematic or photographic styles such as Hong Kong nostalgic cinema and noir. 
We use GPT Image 2.0~\cite{gpt_image_2} to synthesize candidate reference images for each style. Human annotators then review the candidates for stylistic fidelity and visual quality. These references test whether models can apply the depicted visual style to new content without copying the specific subjects or scenes in the reference image.

\paragraph{Structure Reference.}
We construct structure references from source videos using the corresponding procedures in Sec.~\ref{sec:appendix_omni_r2v_construction}.
\begin{itemize}
    \item \textit{Line Art.} We apply frame-wise line extraction, using the same extractor throughout each clip to maintain a consistent representation. These references test whether models can follow the changing contours, poses, and spatial layout throughout a clip while generating appearance details absent from the line maps.
    \item \textit{Rough Storyboard.} We extract temporally ordered keyframes and convert them into sketches that preserve the key subjects, spatial layouts, and actions. These references test whether models can connect sparse structural cues into continuous motion while respecting the key poses, compositions, and their temporal order.
    \item \textit{Greybox.} We convert the initial frame into a greybox representation and use it together with the source video's depth sequence to generate a temporally coherent greybox video. These references test whether models can preserve scene geometry and spatial relations while completing the materials, textures, and other appearance details omitted from the greybox representation.
\end{itemize}

\paragraph{Narrative reference.}
We construct narrative references in three forms: story videos, multi-panel storyboards, and preceding shots.
\begin{itemize}
    \item \textit{Story.} We select existing videos and use them directly as narrative references. These cases test whether models can extract event order and character relationships from a video and preserve the underlying story logic while adapting its visual realization as instructed. 
    \item \textit{Multi-panel Storyboard.} We use Gemini-3.1-Pro~\cite{gemini_3_1_pro} to select 4--9 representative frames from a source video, with the panel count determined by its narrative density. The selected frames are arranged into a multi-panel image with panel indices or an explicitly specified reading order. For a subset of samples, we further use GPT Image 2.0~\cite{gpt_image_2} to convert these images into rough outline sketches or hand-drawn-style storyboards, preserving the basic subject shapes, spatial layouts, and panel sequence while simplifying visual details. The remaining samples retain the original video frames. Human annotators verify that the panels clearly convey the intended story progression, follow an unambiguous temporal order, remain visually legible, and contain no major generation artifacts. It tests whether models can infer event order and generate continuous action from storyboard panels with different levels of visual detail.
    \item \textit{Preceding-shot.} We select clips from existing videos as preceding-shot references for next-shot continuation. The selected cases cover continuations both with and without new characters, objects, or other content introduced in the subsequent shot. These cases test whether models can maintain continuity with the preceding shot while following instructions for subsequent events, including the introduction of new characters or objects.
\end{itemize}

\paragraph{Multi-Content Reference.}
Multi-content references comprise two tasks: entity combination and compositional grounding, with each sample containing 2--5 reference images.
\begin{itemize}
    \item \textit{Entity Combination.} We construct reference combinations such as character--scene, character--object, and character--object--scene. Character references follow the content-reference construction procedures described above, including cross-pair identity matching and generated multi-view images. Object and scene references are extracted from source segments and evaluated by a VLM to ensure that the corresponding objects or scenes are clearly visible. We then use GPT Image 2.0~\cite{gpt_image_2} for inverse construction to produce isolated object images or background-only images. These combinations test whether models can integrate separately referenced characters, objects, and scenes into a coherent video while preserving their identities and realizing the instructed relationships.
    \item \textit{Compositional Grounding.} Compositional-grounding references are constructed by first extracting characters or objects from multiple source videos and then using GPT Image 2.0 to combine multiple characters or multiple objects into a shared reference image. Each target is identified by a distinctive visual attribute or spatial position, allowing the instruction to select and combine specific entities from one or more composite reference images. These cases test whether models can select the intended entities among distractors and bind each selected entity to its instructed role without introducing unselected entities.
\end{itemize}

\paragraph{Cross-Aspect Reference.}
Cross-aspect references comprise four categories: content--motion, content--style, content--structure, and content--narrative. 
\begin{itemize}
    \item \textit{Content--Motion.} Content--motion references are constructed by manually pairing independently curated content references with motion references. This pairing tests whether models can apply the referenced action to the specified content.
    \item \textit{Content--Style}. Content--style references are constructed by manually pairing independently curated content references with style references. This pairing tests whether models can preserve the identities specified by the content references while adopting the style reference’s visual characteristics without copying its semantic content.
    \item \textit{Content--Structure.} We combine content references with either line-art or rough-storyboard references. Structure references follow the corresponding construction procedures in Sec.~\ref{sec:appendix_omni_r2v_construction}, while content references are obtained through cross-segment identity matching. These cases test whether models can place the referenced entities into the poses and spatial arrangements specified by the structural reference while preserving their distinctive appearance.
    \item \textit{Content--Narrative.} Content--narrative references comprise two subtasks: content--story reference and content--multi-panel storyboard. Content--story reference cases are constructed by manually matching existing story videos with content references based on the characters and scenes required by their narratives. Content--multi-panel storyboard cases are constructed by extracting multi-panel storyboards from existing videos and using GPT Image 2.0 to convert them into simplified outline representations that retain the key subjects, spatial layouts, and action sequence. The instruction explicitly specifies the role of each reference.
\end{itemize}

\subsection{Detailed Benchmark Composition}

\label{app:benchmark_composition}
Table~\ref{tab:benchmark_composition} summarizes the detailed composition of OmniVBench across individual sub-tasks. The benchmark contains 813 evaluation cases in total, with cases distributed across both single-reference and multi-reference tasks to provide broad coverage of the R2V task space.

\begin{table}[t]
\centering
\caption{
Detailed composition of OmniVBench.
We report the number of evaluation cases for each sub-task.
}
\label{tab:benchmark_composition}

\setlength{\tabcolsep}{10pt}
\renewcommand{\arraystretch}{0.9}

\begin{tabular}{llr}
\toprule
\textbf{Task Family} & \textbf{Sub-Task} & \textbf{\# Cases} \\
\midrule

\multirow{3}{*}{Content}
    & Object    & 33 \\
    & Character & 54 \\
    & Scene     & 25 \\
\cmidrule(lr){2-3}
    & \textbf{Subtotal} & \textbf{112} \\

\midrule
\multirow{2}{*}{Motion}
    & Action        & 47 \\
    & Camera Motion & 48 \\
\cmidrule(lr){2-3}
    & \textbf{Subtotal} & \textbf{95} \\

\midrule
Style
    & Style & \textbf{43} \\

\midrule
\multirow{3}{*}{Structure}
    & Greybox          & 19 \\
    & Line Art         & 44 \\
    & Rough Storyboard & 45 \\
\cmidrule(lr){2-3}
    & \textbf{Subtotal} & \textbf{108} \\

\midrule
\multirow{3}{*}{Narrative}
    & Multi-Panel Storyboard & 45 \\
    & Story                  & 50 \\
    & Preceding-Shot         & 25 \\
\cmidrule(lr){2-3}
    & \textbf{Subtotal} & \textbf{120} \\

\midrule

\multirow{2}{*}{Multiple Refs.}
    & Multi-Content & 112 \\
    & Cross-Aspect  & 223 \\

\cmidrule(lr){2-3}
    & \textbf{Subtotal} & \textbf{335} \\
\midrule
\multicolumn{2}{l}{\textbf{Total}} & \textbf{813} \\
\bottomrule

\end{tabular}
\end{table}

% Include this file after \appendix in the main manuscript.
\subsection{Evaluation Capability and Criteria}
\label{app:l2_criteria}

The Table~\ref{tab:rf_l2_criteria} and Table~\ref{tab:ir_l2_criteria} list the fixed criteria associated with each L2 sub-dimensions.
For RF and IR, only criteria applicable to a given sample are activated and instantiated as checklist questions. VQ uses the listed automatic evaluation components directly (Table~\ref{tab:vq_l2_criteria}).

\begin{table*}[t]
\centering
\caption{Reference Fidelity (RF) L2 sub-dimensions and evaluation criteria.}
\label{tab:rf_l2_criteria}
\small
\begin{tabular}{p{0.22\textwidth} p{0.68\textwidth}}
\hline
\textbf{L2 Sub-dimensions} & \textbf{Evaluation criteria} \\
\hline
Content Fidelity & Entity/Scene Identity Fidelity; Surface Fidelity; Part Fidelity \\
Structure Fidelity & Structural Alignment; Constraint-Compatible Completion \\
Motion Fidelity & Action Fidelity; Camera Motion Fidelity \\
Style Fidelity & Style Fidelity \\
Narrative Fidelity & Story Structure Fidelity; Continuation State Fidelity \\
\hline
\end{tabular}
\end{table*}

\begin{table*}[t]
\centering
\caption{Instruction Realization (IR) L2 sub-dimensions and evaluation criteria.}
\label{tab:ir_l2_criteria}
\small
\begin{tabular}{p{0.36\textwidth} p{0.58\textwidth}}
\hline
\textbf{L2 Sub-dimensions} & \textbf{Evaluation Criteria} \\
\hline
Reference-Factor Disentanglement and Routing &
Factor--Carrier Separation; Nuisance/Artifact Suppression;
Preserve/Change Factor Partition; Multi-View Identity Unification;
Condition--Slot Correspondence \\

Target Compliance &
Entity Inventory Realization; Attribute/State Realization;
Action Realization; Spatial/Quantity Realization;
Scene/Temporal Setting Realization; Inter-Entity Interaction;
Entity--Object Manipulation; Camera Realization;
Style Realization; Narrative Progression \\
\hline
\end{tabular}
\end{table*}

\begin{table*}[t]
\centering
\caption{Video Quality (VQ) L2 sub-dimensions and automatic evaluation components.}
\label{tab:vq_l2_criteria}
\small
\begin{tabular}{p{0.28\textwidth} p{0.62\textwidth}}
\hline
\textbf{L2 Sub-dimensions} & \textbf{Evaluation component} \\
\hline
Technical Quality & Technical branch of DOVER++~\cite{dover} \\
Aesthetic Quality & Aesthetic Predictor V2.5~\cite{Aes_predictor_v2.5} \\
Physical Plausibility & Coherence/Physics dimension of UnifiedReward 2.0~\cite{unifiedreward} \\
\hline
\end{tabular}
\end{table*}

\paragraph{Criterion activation.}
We do not evaluate every criterion on every case. For each sample, a VLM-based criterion activator takes the instruction and the reference inventory as input and selects, from the full candidate list, only those criteria that the sample genuinely involves; the prompt used for this step is given below. This keeps irrelevant criteria from diluting the score while allowing the applicable set to vary across samples. The complete candidate list substituted into CRITERION CANDIDATES is listed after the prompt.

\paragraph{Checklist instantiation and scoring.}
Each activated criterion is then instantiated into a sample-specific checklist question, which the judge answers from the presented evidence (Sec.~\ref{app:evidence_presentation}) using the rubrics below: Reference Fidelity is scored on a 1–5 scale and Instruction Realization on a 1–3 scale, since the former judges graded degrees of correspondence whereas the latter judges whether an instructed operation was realized at all. Video Quality does not use checklists and is computed directly from the automatic components in Table~\ref{tab:vq_l2_criteria}.

\begin{promptlisting}{Criterion activation prompt}
You are the criterion activator of an evaluation protocol. Given the INSTRUCTION and the REFERENCE INVENTORY of a reference-to-video sample, select from the criterion candidates below those that this sample genuinely involves and should therefore be evaluated on.

[CRITERION CANDIDATES]
{{CRITERION_CANDIDATES}}

[ACTIVATION RULES - must be followed]
A. Activate only the criteria this sample (instruction + references) genuinely involves; never activate one that does not apply.
B. RF (Reference Fidelity) is activated only when the references DO provide that kind of factor AND the instruction asks to preserve it:
   person/object/scene identity -> RF-C01; material, pattern, colour -> RF-C02; local parts -> RF-C03;
   line art / grey box / storyboard structure -> RF-S01, completion of sparse structural gaps -> RF-S02;
   action reference -> RF-M01; camera-motion reference -> RF-M02; style reference -> RF-P01 (fidelity of overall style, medium, colour tone, lighting);
   story structure, key events, character relations, event order and causality -> RF-N01 (event order and causality are covered by this single criterion);
   continuation of the preceding state -> RF-N02.
C. IR-D (disentanglement and routing) is activated when the sample requires ``taking what should be taken from the reference, dropping what should be dropped, and routing multiple conditions correctly'':
   the factor must be separated from its original carrier (e.g. borrow the action but not the identity, borrow the style but not the content) -> IR-D01;
   the reference carries panel lines, draft lines, subtitles or numbering that must be filtered out -> IR-D02;
   [IR-D03 STRICT CONDITION - must be followed] Activate only when the instruction text EXPLICITLY states that some specific factor of the reference must be modified, replaced or removed, e.g. ''change the outfit to a white suit'', ''change the background to a plaza'', put the clothes of Reference Image 1 on the person in Reference Image 2'', ''remove the text in the reference''; only then does a preserve-versus-change partition exist. DO NOT activate when the instruction merely says ''use the character/scene from the reference + describe a new shot/action/scene'' without stating what about the reference should change (such cases are covered by RF for fidelity and by IR-E for realization of the new target, and do not need IR-D03); also do not treat a conversion inherent to the task itself (line art -> coloured, grey box -> textured, draft -> finished shot) as an instruction-requested modification and activate IR-D03 for it.
   multiple viewpoint images of the same entity must be unified into one identity -> IR-D04;
   (only when two or more conditions must be integrated:) multiple references / multiple targets / grounding such as ''use Reference Image 1 on the left and Reference Image 2 on the right'', i.e. pairing conditions to the correct slots -> IR-D05. Do not activate IR-D05 for a single reference with a single target.
D. IR-E (target compliance) is activated only when the TEXT ADDITIONALLY requires some target (i.e. the content is not provided by the references alone):
   added/replaced subject, specified count -> IR-E01; changed colour, material, size, clothing or state -> IR-E02; specified single-subject action -> IR-E03;
   specified position, direction, arrangement, relative distance -> IR-E04 (count belongs to IR-E01 and is no longer a reason to activate this one); specified scene, time, weather or era -> IR-E05; multi-subject interaction -> IR-E06;
   subject-object manipulation (take / use / wear / push / open / ride) -> IR-E07; camera additionally required by the text -> IR-E08;
   style, medium or lighting additionally required by the text -> IR-E09; continuation that advances a new event -> IR-E10.
   Whether a relation holds between the correct objects in multi-subject settings is not a separate criterion: it is carried by the slot correspondence of IR-D05, the inter-subject interaction of IR-E06 and the story structure of RF-N01; listing it separately would double-count.
E. Deduplication: for one and the same action, keep only the most specific among IR-E03 / IR-E06 / IR-E07;
   a factor provided purely by the references goes to RF only and is not repeated in IR-E.

Output exactly one JSON object in this format:
{''RF'': [''RF-C01'', ...], ''I'': [''IR-D01'', ...], ''why'': {''RF-C01'': ''one-sentence justification'', ...}}
Put only criterion IDs in RF/I (they must come from the candidates above); ''why'' gives one sentence of justification per activated criterion. Do not include criteria you did not activate. Output no extra text.



CRITERION_CANDIDATES:

## L1=RF  Reference Fidelity
### RF-C - Content Fidelity
- RF-C01 (Entity / Scene Identity Fidelity): Are the identity and overall form of the subject in the video consistent with the reference? If a scene is involved, are its specific environment and key elements consistent with the reference?
- RF-C02 (Surface Fidelity): Are the materials, textures, patterns and colours designated for reproduction consistent with the reference?
- RF-C03 (Part Fidelity): Are the local parts designated for reproduction consistent with the reference in shape, appearance and detail?
### RF-S - Structure Fidelity
- RF-S01 (Structural Alignment): Does the output follow the geometry, contours, composition, poses, anchors or spatial relations provided by the reference?
- RF-S02 (Constraint-Compatible Completion): Without breaking the structural constraints, are the regions left unspecified by the reference, the pose changes and the motion between anchors completed plausibly?
### RF-M - Motion Fidelity
- RF-M01 (Action Fidelity): Are the key poses, trajectory, amplitude, rhythm, stages and ordering of the subject's action consistent with the reference?
- RF-M02 (Camera Motion Fidelity): Are the type, direction, path and speed of the camera movement, and the ordering of compound movements, consistent with the reference?
### RF-P - Style Fidelity
- RF-P01 (Style Fidelity): Are the overall style, medium, mood, colour tone and lighting of the output consistent with the reference?
### RF-N - Narrative Fidelity
- RF-N01 (Story Structure Fidelity): Are the character relations, key events, event ordering and causal relations specified by the reference reproduced, without semantic substitution or broken relations (visual restructuring requested by the instruction is allowed)?
- RF-N02 (Continuation State Fidelity): Does the next shot continue the characters, objects, environment and narrative state at the end of the reference?
## L1=IR  Instruction Realization
### IR-D - Reference Factor Disentanglement & Routing
- IR-D01 (Factor-Carrier Separation): Is the target factor disentangled from its original reference carrier, without wrongly copying the identity, scene, action semantics, composition or context that carried it?
- IR-D02 (Nuisance / Artifact Suppression): Are carrier elements in the reference - panel lines, draft lines, numbering, subtitles, logos, UI, borders, timestamps, low-resolution noise - all filtered out and absent from the output?
- IR-D03 (Preserve/Change Factor Partition): Are the factors to be faithfully reproduced and the factors to be modified correctly separated - nothing that should change treated as a fidelity item, and nothing that should be preserved damaged?
- IR-D04 (Multi-View Identity Unification): Are multiple reference views of the same entity recognised and merged into one identity, without splitting different viewpoints into separate entities? (Given correct recognition, the per-view fidelity of that identity is still judged by RF-C01.)
- IR-D05 (Condition-Slot Correspondence): Is every reference factor and textual condition assigned to the target/slot specified by the instruction, with no condition mismatch?
### IR-E - Target Compliance
- IR-E01 (Entity Inventory Realization): Do the characters, objects, categories and counts required by the text appear correctly (including added entities, replaced subjects and designated objects)? (Attribute detail and reference identity fidelity are not judged here.)
- IR-E02 (Attribute / State Realization): Are the colours, materials, sizes, local attributes, clothing, states or attribute replacements required by the text correctly realized in the output? (The preserve/modify boundary of reference attributes is not judged here.)
- IR-E03 (Action Realization): Does the single-subject action, pose change or action semantics required by the text - and not fully determined by a motion reference - occur correctly? (Fidelity to a motion reference is not judged here.)
- IR-E04 (Spatial / Quantity Realization): Are the positions, directions, counts, arrangement, relative distances or spatial relations required by the text correctly realized? (Fidelity to a structural reference is not judged here.)
- IR-E05 (Scene / Temporal Setting Realization): Are the scene, environment, time, weather, era or location conditions required by the text correctly realized? (Fidelity to a scene reference is not judged here.)
- IR-E06 (Inter-Entity Interaction): Does the multi-subject interaction, contact, confrontation, cooperation or relational action required by the text occur correctly? (Participant identity and its correspondence to the source are judged by IR-D05.)
- IR-E07 (Entity-Object Manipulation): Does the subject-object manipulation required by the text (taking, using, wearing, pushing, opening, riding, etc.) occur correctly? (Fidelity of the object's appearance is not judged here.)
- IR-E08 (Camera Realization): Are the viewpoint, shot scale, composition, camera movement or shot transitions additionally required by the text correctly realized? (Fidelity to a camera-motion reference is judged by RF-M02.)
- IR-E09 (Style Realization): Are the style, medium, lighting, colour tone, mood or rendering approach additionally required by the text correctly realized? (Fidelity to a style reference is judged by RF-P01.)
- IR-E10 (Narrative Progression): Does the generated content realize the new event, state change, causal advance or story development required by the text, rather than merely repeating the end of the reference or statically reproducing the reference state?
\end{promptlisting}

\begin{promptlisting}{Scoring rubric for Reference Fidelity (1–5 scale)}
Score each sub-question 1-5 from visual evidence only. Judge ONLY the signal explicitly asked about. First identify that question's CORE requirements; unrelated differences do not cost points. Do not penalize general video quality unless it prevents the asked reference factor from being reliably observed or preserved. Evaluate a signal at its RELEVANT moments or opportunities. Do not require an action, event, transition or pose to appear in every frame merely to earn 5.
Apply the scale in this order: first decide whether every CORE requirement is correct; if not, decide whether the SPECIFIC intended signal is still recognisable, only its broad category is related, or no meaningful correspondence remains.
- 5 = complete and clean: every core requirement and every clearly observable secondary aspect of the asked signal matches. The signal is shown with enough evidence at its relevant moments, and no specific visible deviation can be identified. Do not give 5 merely because the overall impression is correct.
- 4 = complete with a minor visible flaw: every core requirement matches, but at least one specific non-core deviation can be identified, such as a small difference in degree, a secondary-detail mismatch or a brief local drift. It must not remove, replace, reverse or materially alter any core requirement.
- 3 = specifically recognisable but materially incomplete: enough defining evidence remains to identify the specific intended signal, but at least one core requirement is visibly missing, wrong, substantially different or unstable. Also use 3 when the core signal appears correct but the output does not provide enough evidence to verify it reliably. The match must go beyond category or general resemblance.
- 2 = only loosely related: some positive correspondence is visible, but most defining requirements are missing or wrong. The output matches only the broad category, theme, action family, visual family or rough impression; the specific referenced signal cannot be established reliably.
- 1 = no meaningful correspondence: the asked signal is absent, unrelated, replaced, contradicted or reversed. Unlike score 2, there is not enough positive evidence even for a weak attempted match.
ALLOWED TRANSFORMATIONS: Discount a difference only when the instruction explicitly requests it or it is inherent to the required carrier conversion, and only when the sub-question does not ask about that factor. Examples include line art or a grey-box becoming a finished render, still panels becoming continuous motion, or an explicitly requested change of identity, appearance, scene or style. Re-framing, re-staging, camera changes, duration and pacing are not blanket exemptions: ignore them only when they are explicitly permitted by the instruction or the provided evaluation context and they do not damage the signal being scored. A permitted transformation is not itself an error, but any resulting loss of the asked signal still counts.
VISIBILITY AND EVIDENCE: Do not invent a mismatch in detail that the intended output shot scale cannot resolve, and discount apparent changes caused only by viewpoint, expression, lighting, motion blur or ordinary resolution limits. However, output-caused concealment is not proof of fidelity: if the output's own framing, cropping, occlusion, blur or brevity prevents a required factor from ever being verified, it cannot receive 4 or 5. If the required subject or event is effectively absent, score 1 or 2 rather than treating it as unobservable.
IDENTITY ('the same person / object / scene'): judge instance sameness, not category likeness. Generic traits prove little. Compare individuating structure: facial geometry and feature proportions, hairline and unique marks; object shape, proportions and workmanship; scene-specific layout and landmarks. 5 = clearly the same instance with all verifiable defining structure matching; 4 = clearly the same with one minor defining deviation; 3 = recognisably derived from the same instance but with one substantial or several visible deviations; 2 = only a similar member of the same category or too weakly shown to establish sameness; 1 = different or absent.
DETAIL / SURFACE / PART CONSISTENCY: Weight differences by importance, not raw count. One wrong defining part may be a core failure, while several incidental details may be minor. Judge the coverage and severity of visible, resolvable defining features.
MOTION / CAMERA MOTION: Core requirements are the requested stages, direction, trajectory, extent, key poses and ordering. Judge each at the corresponding relevant part of the ordered frames; a correctly completed one-time stage may earn 5. Never invent motion between frames, and judge absolute timing only when the supplied frame note says it is comparable.
STRUCTURE / COMPOSITION: Core requirements are the anchors actually named by the question, such as outline, pose, placement, proportion, shot scale, composition and spatial relations. A difference in one of those named factors is not re-staging to discount.
COMPLETION BETWEEN ANCHORS: Judge whether the unspecified intervals are filled continuously, naturally and without breaking the given anchors. Do not demand pixel alignment to an anchor during an interval that is supposed to introduce motion or new content.
NARRATIVE: Core requirements are the named events, roles, relations, causal links and order. Score semantic preservation rather than visual identity, setting, shot choice or absolute pace unless the question explicitly names those factors.
STYLE: Judge the named overall visual language, medium, palette, material rendering, lighting or atmosphere. Ignore changed content and composition unless they are part of the style signal explicitly asked about.
\end{promptlisting}

\begin{promptlisting}{Scoring rubric for Instruction Realization (1–3 scale)}
Score each sub-question 1-3 from visual evidence only. Judge only whether the specified instruction, operation or factor-target assignment is correctly realized.
- 3 = fully realized: all core requirements are satisfied, every specified factor is applied to the correct target, and prohibited content is absent at all observable relevant moments.
- 2 = partially realized: the intended result is clearly present and at least one core requirement is satisfied, but another requirement is incomplete, locally misplaced, unstable or intermittently violated.
- 1 = not realized: the intended result is absent or fundamentally wrong, no core requirement is reliably satisfied, the main factor is assigned to the wrong target, or prohibited content substantially remains.
Do not judge general video quality. Do not score the degree of visual similarity unless it is necessary to determine whether the requested factor was selected, rejected or assigned to the correct target.
\end{promptlisting}

\subsection{Evidence Presentation}
\label{app:evidence_presentation}

Evidence presentation specifies how the reference and generated videos are shown to the judge during RF and IR evaluation. Its purpose is to provide consistent visual evidence for every model output while preserving the temporal information relevant to the task. We therefore organize the evidence according to the role that each reference plays in the task, as described below.

\paragraph{Non-temporally aligned references.}
For non-temporally aligned references, such as content, presentation/style, and multi-panel references, we uniformly sample frames from the generated video at 2 FPS, with a minimum of 8 and a maximum of 20 sampled frames. The sampled generated frames are presented jointly with the corresponding reference inputs to the VLM judge, enabling direct comparison of the referenced visual factors and their realization in the generated video.

\paragraph{Temporally aligned references.}
For motion references and temporally structured references such as greybox and line art, we sample ten reference–output frame pairs at matched normalized temporal positions and present them in temporal order to the judge.

\paragraph{Rough storyboards.}
For rough storyboards, we retain ordered, non-redundant anchor frames rather than treating every storyboard cell as an exact frame-level target. The anchors are compared with twelve temporally ordered frames from the generated video.
This preserves the intended event order, scene transitions, and major compositional changes while allowing the generated video to interpolate between the storyboard's sparse visual instructions.

\paragraph{Continuation references.}
For continuation tasks, frames from the preceding shot are presented before frames from the generated continuation. This makes the boundary between the provided context and generated content explicit, so the VLM judge can assess both state continuity and the requested subsequent development.

\subsection{Human Evaluation Protocol}

\begin{table}[t]
    \centering
    \caption{
        Five-point rating criteria for human evaluation of
        Instruction Realization (IR) and Reference Fidelity (RF).
        Only requirements applicable to each sample are assessed.
    }
    \label{tab:human_evaluation_rubric}
    \small
    \setlength{\tabcolsep}{5pt}
    \renewcommand{\arraystretch}{1.18}
    \begin{tabular}{
        @{}
        c
        p{0.43\linewidth}
        p{0.43\linewidth}
        @{}
    }
        \toprule
        \textbf{Score}
        & \textbf{Instruction Realization}
        & \textbf{Reference Fidelity} \\
        \midrule

        1
        & The instruction is not executed, or the output is
          unrelated to the requested task.
        & The required reference information is absent, or
          the output has no meaningful correspondence with
          the reference. \\
        \addlinespace

        2
        & The output shows an attempt to follow the instruction
          but largely fails or introduces incorrect entities.
        & The required reference information is only weakly
          preserved, with substantial deviations that make
          the correspondence difficult to recognize. \\
        \addlinespace

        3
        & The instruction is partially fulfilled, but key
          requirements are missing or incorrectly realized.
        & The required reference information is recognizable
          but only partially preserved, with clear
          inconsistencies or errors. \\
        \addlinespace

        4
        & The main requirements are fulfilled, with minor
          deviations in details.
        & The required reference information is largely
          preserved, with minor local or temporal
          inconsistencies. \\
        \addlinespace

        5
        & All applicable instruction requirements are fulfilled,
          including specified entities, attributes, positions,
          durations, and quantities.
        & The required reference information is faithfully
          preserved throughout the relevant portions of the
          video, with no substantive inconsistencies. \\

        \bottomrule
    \end{tabular}
\end{table}

Three annotators independently evaluate each generated video using the textual instruction and the corresponding visual references. They assign separate scores for Instruction Realization (IR) and Reference Fidelity (RF) on a five-point scale, using the criteria in Table~\ref{tab:human_evaluation_rubric}.

\subsection{More Results}

\begin{table*}[t]
\centering
\caption{Human evaluation of Omni-R2V Dataset quality across the seven task families. Each entry reports the average Pass rate (\%) of two annotators.}
\label{tab:data_quality}
\small
\setlength{\tabcolsep}{4pt}
\scalebox{0.87}{
\begin{tabular}{lcccccccc}
\hline
\textbf{Criterion} &
\textbf{Content} &
\textbf{Motion} &
\textbf{Style} &
\textbf{Structure} &
\textbf{Narrative} &
\textbf{Multi-Content} &
\textbf{Cross-Aspect} &
\textbf{Avg.} \\
\hline
Reference Usability
& 97.5 & 90.0 & 100.0 & 97.0 & 92.5 & 96.0 & 94.5 & 95.4 \\

Reference--Target Consistency
& 100.0 & 92.5 & 98.0 & 98.5 & 98.5 & 96.0 & 95.5 & 97.0 \\

Instruction Correctness
& 94.5 & 86.5 & 96.5 & 94.5 & 99.0 & 94.0 & 94.5 & 94.2 \\
\hline
Overall
& 97.3 & 89.7 & 98.2 & 96.7 & 96.7 & 95.3 & 94.8 & 95.5 \\
\hline
\end{tabular}}
\end{table*}
\paragraph{Human Evaluation of Data Quality.}
We conduct a human evaluation by randomly sampling approximately 100 training samples from each of the seven task families. Two annotators independently inspect each sample using Pass/Fail judgments along three criteria:
(1) \textit{Reference Usability}, whether the references clearly contain the information required by the task;
(2) \textit{Reference--Target Consistency}, whether the target correctly reflects the reference factors that should be preserved or transferred; and
(3) \textit{Instruction Correctness}, whether the instruction accurately specifies the roles of the references and the intended target requirements.
We report the mean pass rate across the two annotators for each task family and criterion, together with their averages. Disagreements between annotators are resolved through review.

\subsubsection{Qualitative Results}

\begin{figure*}[t]
\centering{\includegraphics[width=0.99\textwidth]{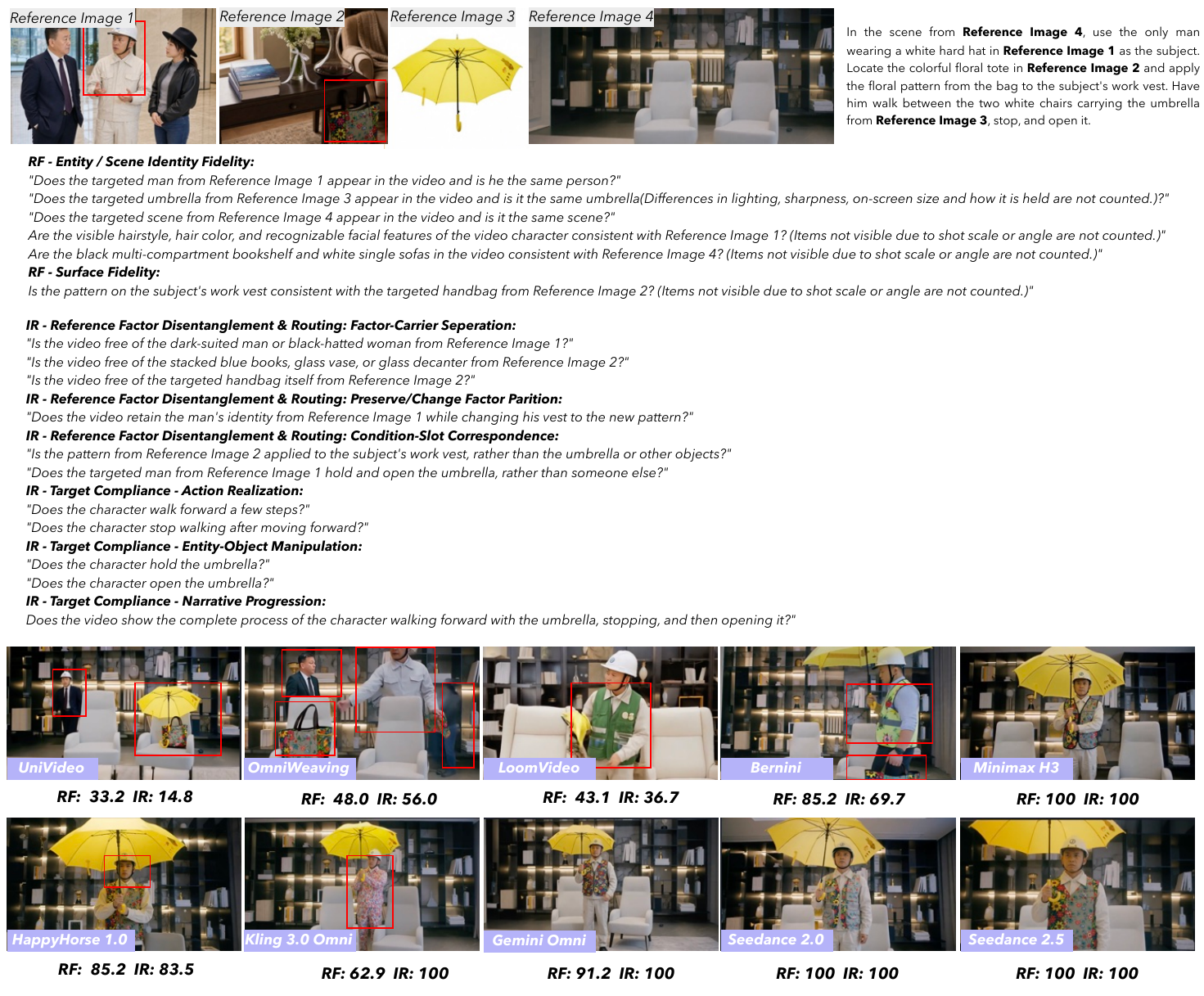}}
\caption{Example of factor-grounded evaluation on a multi-reference case, showing the case-specific RF/IR checklists and corresponding model scores.}
\label{Fig.case_score}
\end{figure*}
\paragraph{Qualitative Analysis of Factor-Grounded Evaluation.}
Fig.~\ref{Fig.case_score} presents a representative multi-reference case illustrating our factor-grounded evaluation. The case-specific RF and IR checklists capture fine-grained differences in reference fidelity, reference-factor disentanglement and routing, and target compliance across models.

\begin{figure*}[t]
\centering{\includegraphics[width=0.99\textwidth]{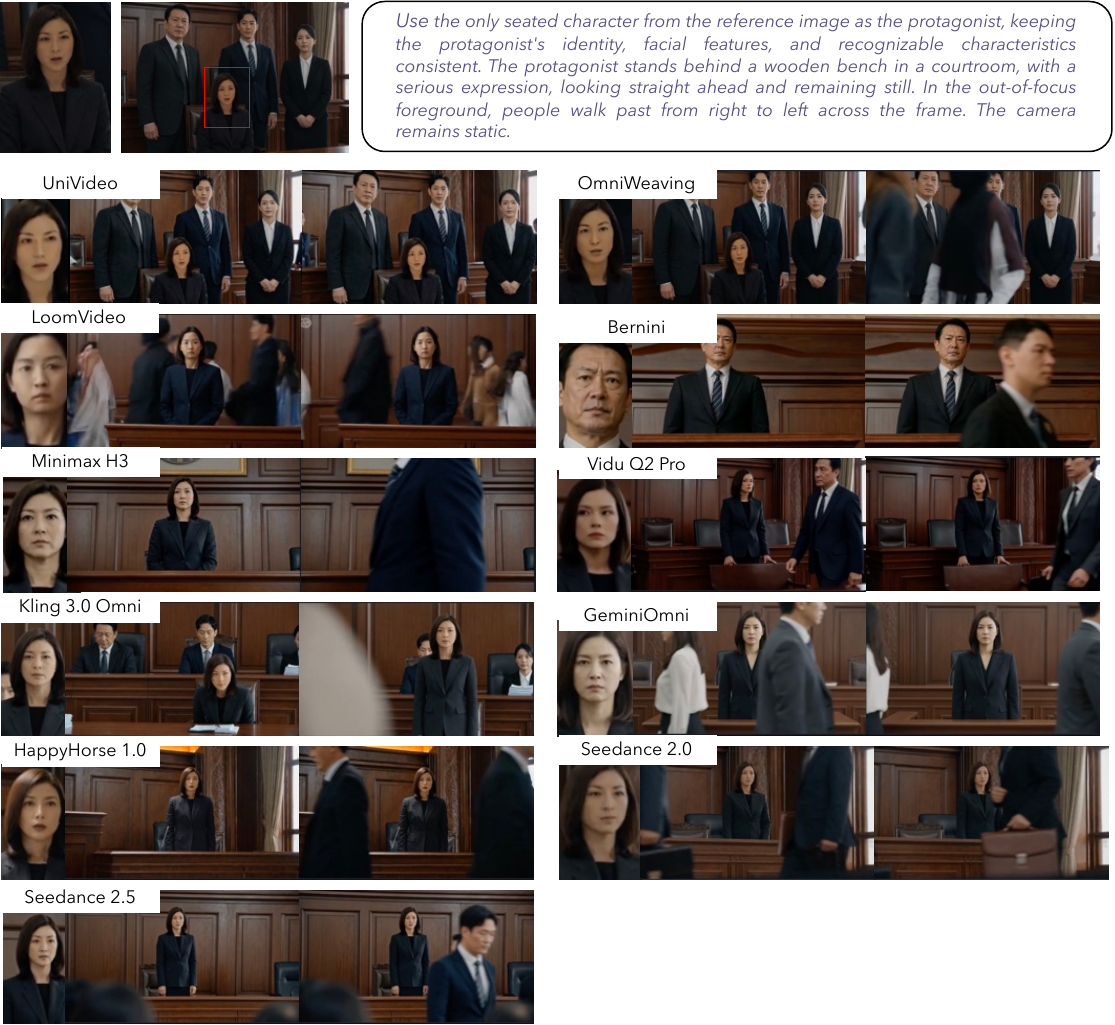}}
\caption{Qualitative comparison on the content reference task. }
\label{Fig.case_content1}
\end{figure*}

\begin{figure*}[t]
\centering{\includegraphics[width=0.99\textwidth]{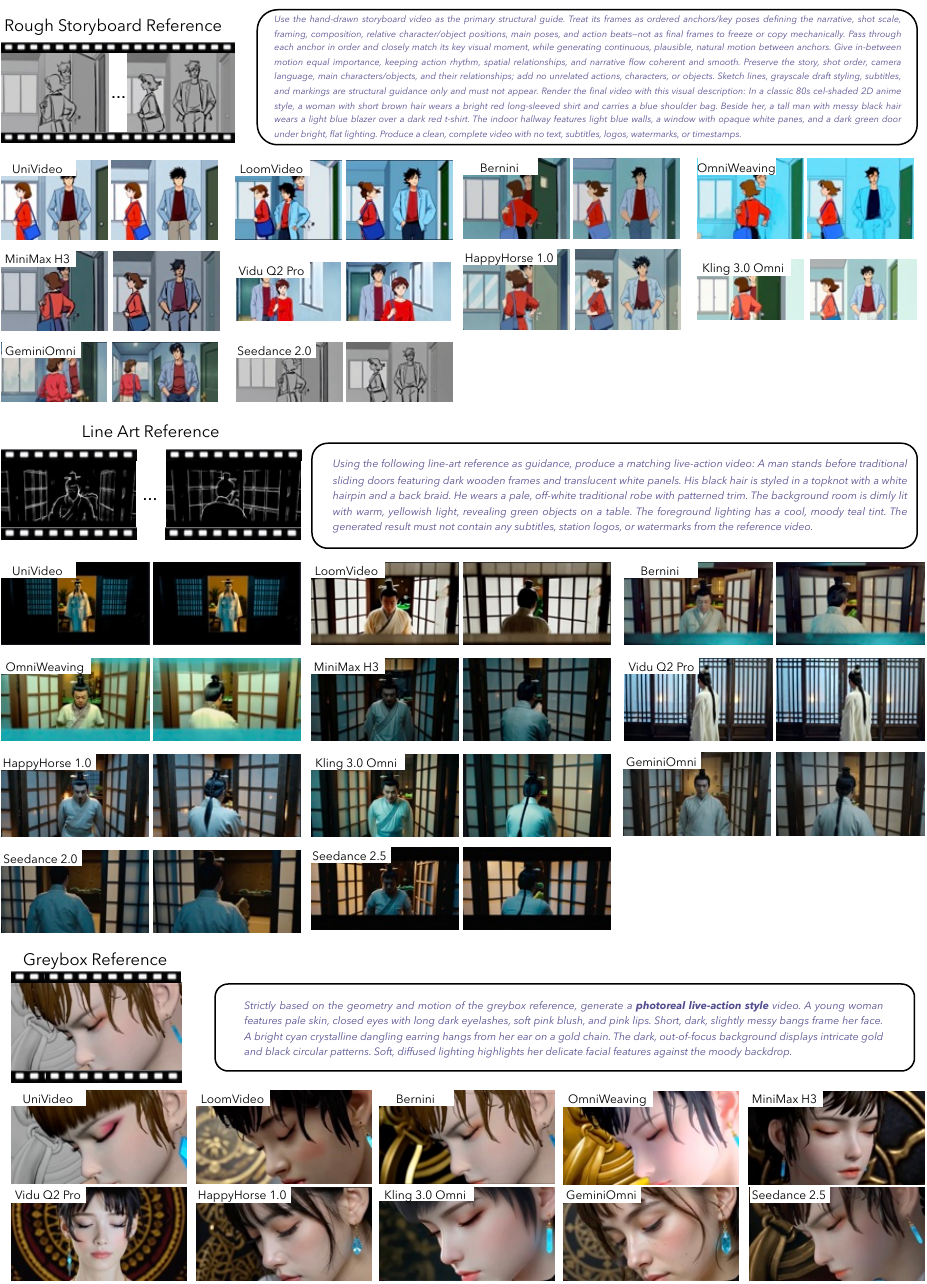}}
\caption{Qualitative comparison on the structure reference task.}
\label{Fig.case_structure}
\end{figure*}

\begin{figure*}[t]
\centering{\includegraphics[width=0.99\textwidth]{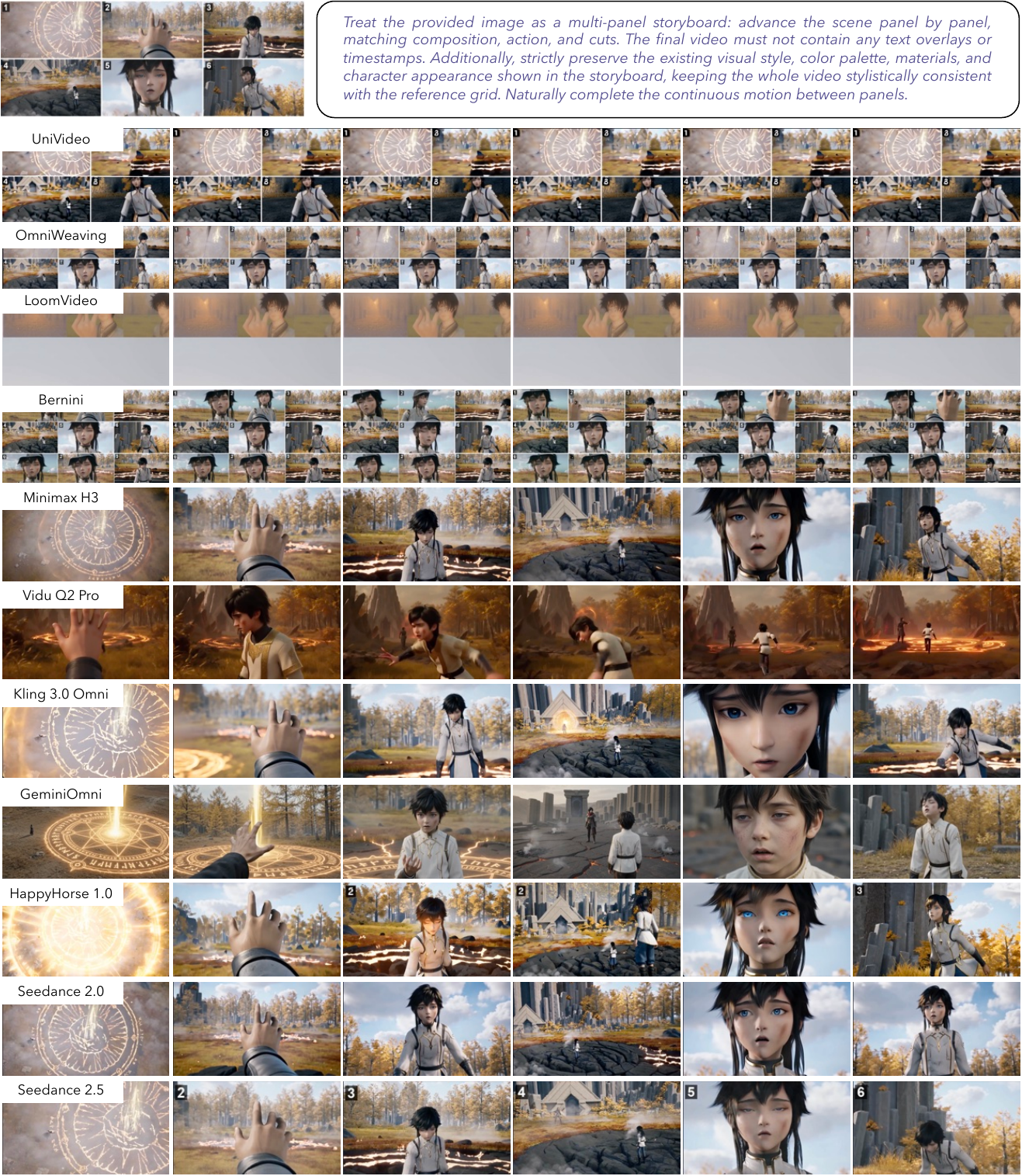}}
\caption{Qualitative comparison on the narrative reference task. }
\label{Fig.case_multipanel1}
\end{figure*}

\begin{figure*}[t]
\centering{\includegraphics[width=0.99\textwidth]{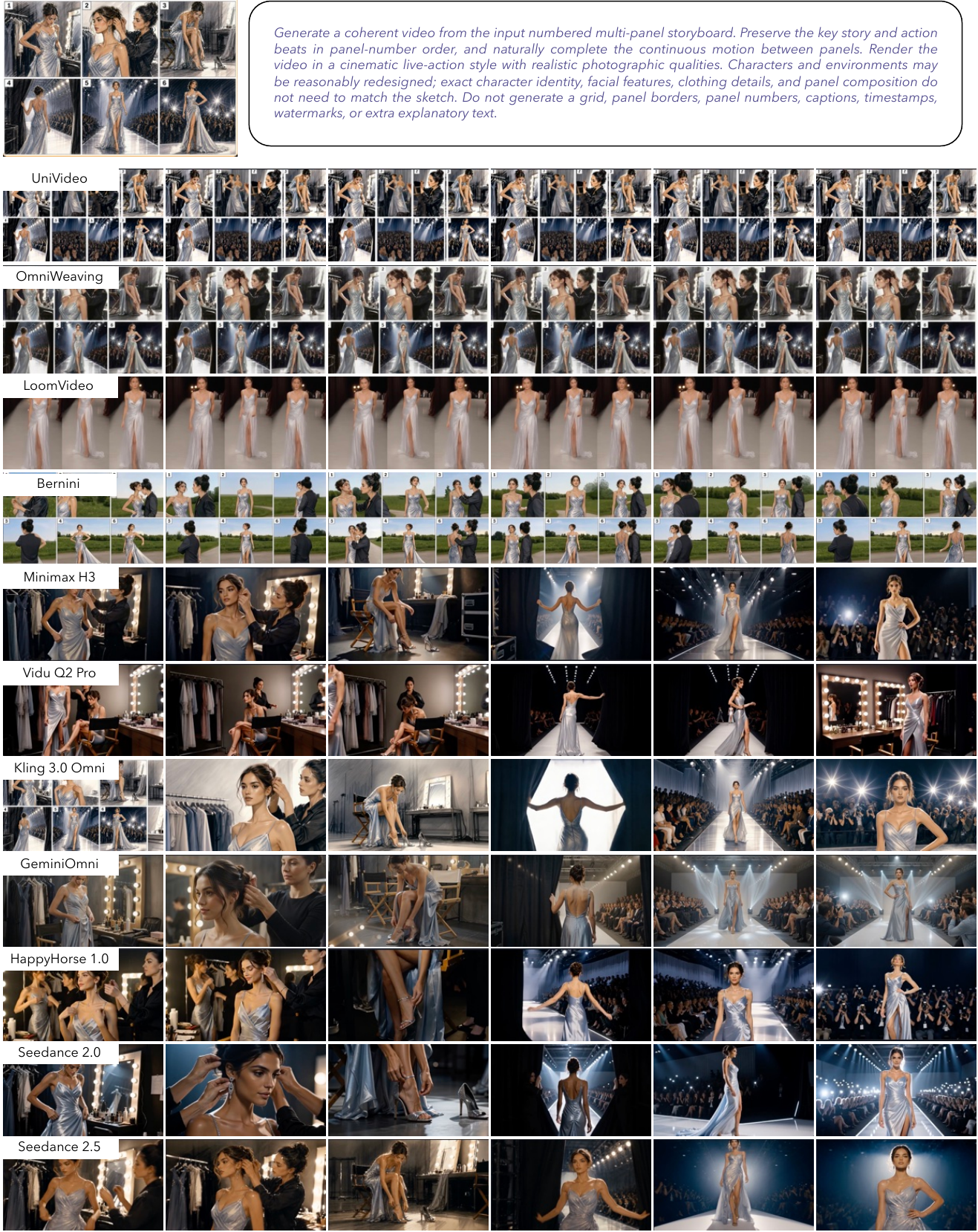}}
\caption{Qualitative comparison on the narrative reference task. }
\label{Fig.case_multipanel2}
\end{figure*}

\begin{figure*}[t]
\centering{\includegraphics[width=0.99\textwidth]{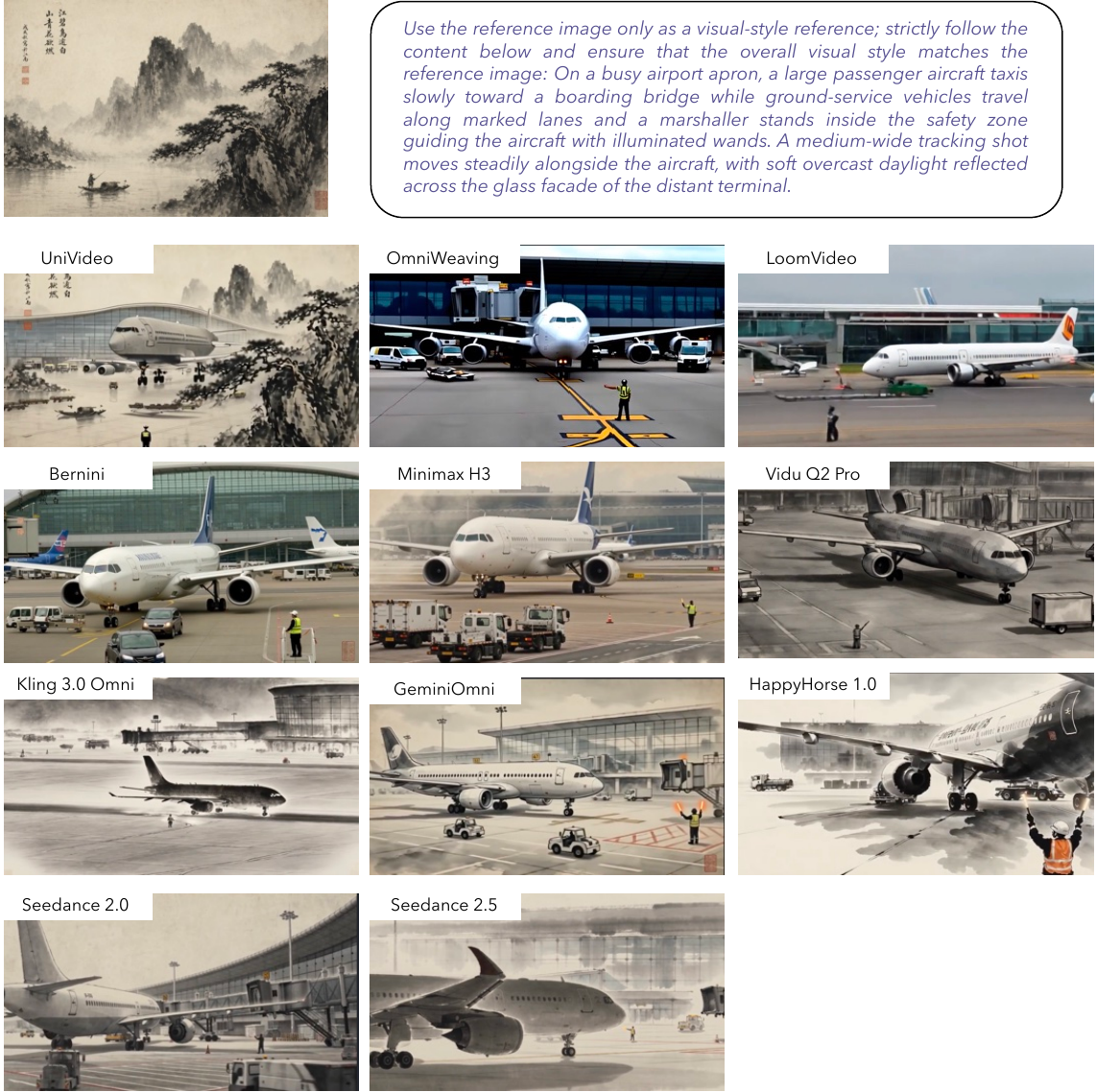}}
\caption{Qualitative comparison on the style reference task. }
\label{Fig.case_style}
\end{figure*}

\begin{figure*}[t]
\centering{\includegraphics[width=0.99\textwidth]{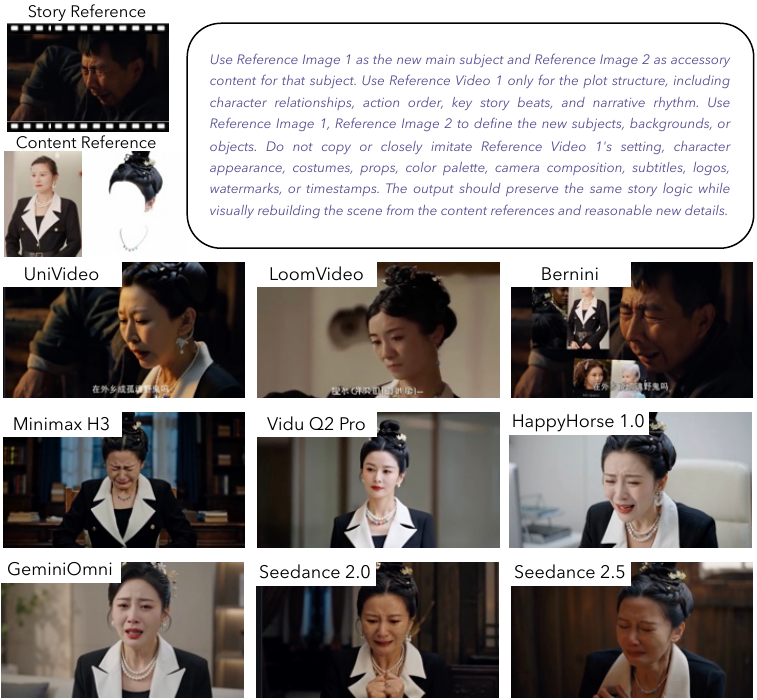}}
\caption{Qualitative comparison on the cross-aspect reference task.}
\label{Fig.case_content_style}
\end{figure*}

\paragraph{Qualitative Comparison.}
Additional qualitative comparisons across content, structure, narrative, style, and cross-aspect reference tasks are provided in
Figs.~\ref{Fig.case_content1}--\ref{Fig.case_content_style}.

\end{document}